\documentclass[journal]{IEEEtran}
\usepackage[utf8]{inputenc}
\usepackage{graphicx}
\usepackage[caption=false,font=footnotesize]{subfig} % subfigures, set up according to IEEEtran documentation
\usepackage[tbtags]{amsmath} % tbtags - IEEE: "Equation numbering should be [...] on line with the last line of an equation"

\usepackage{amsthm} \theoremstyle{remark} \newtheorem{remark}{Remark}
\usepackage{mathrsfs} % Ralph Smith’s Formal Script Font e.g., range R
\usepackage{hyperref} % hyperlinks for urls and references
\usepackage{tcolorbox} % Boxes around subproblem definitions
\usepackage{booktabs} % horizontal rules for tables
\usepackage{xcolor} % Colored text for comments
\usepackage{threeparttable} % Footnotes in tables
\usepackage{needspace}

\newif\ifabridgedsolutions
\abridgedsolutionsfalse
\DeclareSymbolFont{AMSb}{U}{msb}{m}{n}
\def\rot{\textit{\textsf R}}
\def\norm#1{\left\|{#1}\right\|}
\def\abs#1{\left|{#1}\right|}
\def\tr{^T}
\DeclareMathSymbol{\RR}{\mathbin}{AMSb}{"52}
\def\rr#1{\RR^{#1}}
\def\c#1{\mbox{c}_{#1}}
\def\s#1{\mbox{s}_{#1}}
\def\sq{^2}
\def\ma{\left[ \begin{array}}
\def\ema{\end{array}\right]}
\def\cross{\times}

\tcbset{sharp corners,
    toptitle=0.25em,
    bottomtitle=0.25em,
    colback=white,
    enlarge top initially by=0.5em,
    phantom = \phantomsection,
    before = {\par\smallskip\parindent=0pt},
    after={\needspace{2\baselineskip}}}
\newtcolorbox[list inside=toc, list type = subsection]{subsectionbox}[1]{title=\textsc{#1}}

\title{IK-Geo: Unified Robot Inverse Kinematics\\Using Subproblem Decomposition}
\author{Alexander~J.~Elias,~\IEEEmembership{Graduate Student Member,~IEEE}, and John~T.~Wen,~\IEEEmembership{Fellow,~IEEE}% <- Prevent space
\thanks{
\textit{(Corresponding author: Alexander J. Elias.)}

The authors are with the Department of Electrical, Computer, and Systems Engineering, Rensselaer Polytechnic Institute, Troy, NY 12180 USA (e\mbox{-}mail:~eliasa3@rpi.edu; wenj@rpi.edu).}%
}

\begin{document}
\maketitle
\begin{abstract} % No more than 200 words or 1200 characters for T-RO
This paper presents the open-source robot inverse kinematics (IK) solver IK-Geo, the fastest general IK solver based on published literature. In this unifying approach, IK for any 6-DOF all-revolute (6R) manipulator is decomposed into six canonical geometric subproblems solved by intersecting circles with other geometric objects. We present new efficient and singularity-robust solutions to these subproblems using geometric and linear algebra methods. IK-Geo finds all IK solutions including singular solutions and sometimes least-squares solutions by
    solving for subproblem solutions in all cases, including in a continuous and sometimes least-squares sense when a solution does not exist.
Robots are classified into kinematic families based on cases of intersecting or parallel joint axes, and robots in the same family use the same IK algorithm.
6R robots with three intersecting or parallel axes are solved in closed form, and all solutions are found exactly without iteration. Other 6R robots are efficiently solved by searching for zeros of an error function of one or two joint angles. The subproblem and IK solutions are easy to understand, implement, test, and modify, meaning this method is readily ported to new languages and environments.
We connect our geometric method with less efficient but more robust polynomial-based methods: rather than using search, subproblems and error functions may be written in terms of the tangent half-angle of one joint. This results in a system of multivariate polynomial equations from which the univariate polynomial with zeros corresponding to IK solutions is readily derived.
\end{abstract}

\section{Introduction}
\setlength\tabcolsep{3pt} % narrower vertical spacing (default value: 6pt)
\begin{table}[t]
    \centering
    \caption{Classification of All 6R Robot Manipulator Kinematics}
    \begin{threeparttable}
    \begin{tabular}{ l l }
    \toprule
     Robot Kinematic Family & Example\\ \midrule
     Spherical joint                    & Franka Production 3 \cite{franka}, fixed \(q_5\)\\
     \quad and two intersecting axes    & KUKA LBR iiwa 7 R800 \cite{KUKA}, fixed \(q_3\)\\
     \quad and two parallel axes        & ABB IRB 6640 \cite{ABB_IRB_6640}\\
     Three parallel axes                & N/A\tnote{*} \\
     \quad and two intersecting axes    & Universal Robots UR5 \cite{UR5}\\
     \quad and two parallel axes        & N/A\tnote{*} \\
     Two intersecting axes              & Kassow Robots KR810 \cite{kassow}, fixed \(q_7\)\\
     \quad and two intersecting axes    & FANUC CRX-10iA/L \cite{crx}\\
     \quad and two parallel axes        & Kawasaki KJ125 \cite{kawasaki}\\
     Two parallel axes                  & N/A\tnote{*} \\
     \quad and two parallel axes        & N/A\tnote{*} \\
     Two intersecting axes \(k, k{+}2\)\tnote{\textdagger} & ABB YuMi \cite{ABB_YUMI}, fixed \(q_3\) \\
     \quad and two intersecting axes    & RRC K-1207i \cite{RRC}, fixed \(q_6\)\\
     \quad and two parallel axes        & N/A\tnote{*} \\
     General 6R                         & Kassow Robots KR810 \cite{kassow}, fixed \(q_6\)\\
     \bottomrule
    \end{tabular}
    \begin{tablenotes}
        \item[*] No such industrial robot.
        \item[\textdagger] Axes \(k\) and \(k+2\) are nonconsecutive intersecting axes. All other intersecting or parallel cases refer to consecutive axes.
    \end{tablenotes}
    \end{threeparttable}
    \label{tab:robot_examples}
\end{table}
\setlength\tabcolsep{6pt} % return to default
\IEEEPARstart{R}{obotics} applications have always demanded a fast and reliable inverse kinematics (IK) algorithm
	which can ideally return all solutions for a given end effector pose.
Fast IK is important for
	real-time Cartesian control and path planning
	as well as for offline conversion of trajectories from task space to joint space.
Returning all IK solutions is critical for path planning, and
singularity-robust IK algorithms help avoid resorting to joint-space control to switch between IK solution branches.

There are many classes of robot kinematics based on intersecting or parallel axes as shown in Table~\ref{tab:robot_examples}.
Nonconsecutive intersecting joint axes \(k, k+2\) occur when they are both orthogonal to axis \(k+1\) or when the two links have equal lengths, zero offset, and opposite twists, which is a special case of the Bennett criteria \cite{lu2017approximation}.
An important goal for an IK solver is not only to solve the general 6-DOF all-revolute (6R) case but also to use simplifications arising from intersecting or parallel axes to improve computational performance.

Gradient-based solvers,
	such as the popular solver TRAC-IK~\cite{beeson2015trac},
	use the robot Jacobian and can run sufficiently quickly for real-time control.
However, they only return one solution close to the initial guess,
    suffer from performance issues close to robot singularities,
    and may not always return solutions within a given time period,
    especially if there is no close initial guess
    or if there is a high accuracy requirement.
\IEEEpubidadjcol% must be placed in second column of first page

In contrast, analytical IK solvers can return all inverse kinematics solutions
	and can run much faster than gradient-based solvers.
IKFast~\cite{diankov_thesis} is the most widely used analytical IK solver and has been used in large projects such as MoveIt~\cite{moveit} and Tesseract~\cite{tesseract}, as well as smaller standalone projects.
However, IKFast cannot solve all 6R robots, and it depends on symbolic manipulation techniques and exploring possible simplifications,
	meaning code generation and even compilation can take a considerable amount of time.
IKFast is a Python file nearly 10\,000 lines long which generates an even larger C++ file, ranging from about 10\,000 lines to over 100\,000 lines in the case of a UR5 robot.
	This means modifying, optimizing, or debugging by hand is often impractical.
Code must be regenerated for different kinematics parameters even if only link lengths differ,
	and solutions break down at singularities.
IKBT~\cite{zhang2019ikbt} is a newer analytical IK solver similar to IKFast,
	but it is still based on symbolic manipulation and simplification exploration,
	and so it suffers from similar issues.
There has also been a recent resurgence in AI algorithms for IK~\cite{malik2022deep, ames2022ikflow}, but they require extensive tuning and training and do not give precise results. 
Some authors have also found general 6R IK methods based on finding a high-order polynomial in the tangent half angle of of joint \cite{raghavan1990kinematic, raghavan1993inverse, husty2007new}, but these methods are difficult to implement in real-time settings because the methods require performing symbolic algebraic manipulations for each new end effector pose.

\begin{table}[t]
    \centering
    \caption{IK Solution Types and Subproblems Used}
    \begin{threeparttable}
    \begin{tabular}{l l l l l }
        \toprule
        Solution Type & Robot Kinematic Family             & SPs Used         & Section                \\ \midrule
        Closed-form   & Spherical joint            & 1, 2, 5\tnote{*} & \ref{sec:three_intersecting} \\
                      & Three parallel axes                & 1, 3, 6\tnote{*} & \ref{sec:3_parallel}         \\ \midrule
        1D search     & Two intersecting axes              & 1, 5\tnote{*}    & \ref{sec:2_intersecting}     \\
                      & Two parallel axes                  & 1, 6\tnote{*}    & \ref{sec:2_parallel}         \\
                      & Two intersecting axes \(k, k{+}2\) & 1, 5\tnote{*}    & \ref{sec:2_nonconsecutive_intersecting} \\ \midrule
        2D search     & General 6R                         & 1, 5             & \ref{sec:gen_6R}             \\ \bottomrule
    \end{tabular}
    \begin{tablenotes}
        \item[*] Simplifies to Subproblems 1--4 if extra intersecting or parallel axes.
    \end{tablenotes}
    \end{threeparttable}
    \label{tab:robot_kin_families}
\end{table}

In this paper, we present IK-Geo,
	a highly capable analytical and semi-analytical general IK solver for 6R arms
	based on a unifying and simplifying geometric approach called subproblem decomposition.
This method solves for a small number of joint angles at a time
	by exploiting simplifications arising from intersecting or parallel joint axes
	and rewriting the problem as a series of a few canonical geometric subproblems with known solutions.
This results in algorithms which are
	precise, 
	computationally efficient,
	stable,
	and which return all solutions, including singular solutions and sometimes least squares solutions.
IK-Geo is the fastest general solver based on published literature.
For example, in our testing,
    IK for a UR5 robot using subproblem decomposition is more than 40 times faster than IKFast.
 
Our closed-form solutions return all IK branches even for end effector poses outside the workspace, and these solutions are continuous with respect to the desired end effector pose. This helps achieve the goal of having a continuous joint trajectory for any given continuous end effector trajectory, even if it falls outside the robot workspace.
In certain cases, such as when joint axes are collinear, the robot encounters an internal singularity, and a continuum of self-motion is possible. In this case, a subproblem may have a continuum of solutions rather than finitely many solutions.
If redundancy resolution is applied correctly, end effector trajectories that encounter internal singularities can also result in continuous joint trajectories.

Depending on cases of intersecting or parallel joint axes, robot IK is solved in closed form or with a reduced dimension search over one or two joint angles, as shown in Table~\ref{tab:robot_kin_families}.
To the best of our knowledge, all commercially available industrial 6R robots and 7R robots parameterized by some joint angle have intersecting or parallel axes and therefore can be solved in closed form or with 1D search.
The closed-form method applies to any 6R robot with three consecutive intersecting or parallel axes
	(commonly referred to as the Pieper criterion \cite{pieper1969kinematics}),
	and requires only addition, subtraction, multiplication, division, square roots, ATAN2, and (in the case of solving quartic polynomials) cube roots.
Otherwise, searching over just one or two joint angles is much more efficient than searching over all six joint angles as in Jacobian-based methods.
The 1D and 2D search cases also offer a useful graphical output.
To the best of our knowledge, we are the first paper to apply only subproblems to solve IK for
    any general 6R robot with 2D search,
    any 6R robot with any two intersecting or parallel axes with 1D search,
        including when the intersecting axes are nonconsecutive,
    and any 6R robot with three parallel axes in closed form.

We also show how to extend our search solutions to solve IK by finding the zeros of a polynomial,
	and we are the first to make this connection between subproblem decomposition and polynomial methods.
Although the polynomial method is less efficient than search methods
	and resorts to the tangent half angle substitution and symbolic manipulation,
	this method guarantees finding all solutions.
Using the same subproblem decompositions and error function formulations as in the search method,
	we apply the tangent half angle substitution
	which results in a system of three or four multivariate polynomials.
The resultant of this system is a polynomial in one unknown, which may be solved to arbitrary precision.
Once one or two joint angles are found, the remaining joint angles are found in closed form.
We share insights on improving computational performance by
	efficiently parameterizing end effector orientation
	and by using rational approximations of kinematic parameters and end effector pose.
Unlike other similar polynomial methods,
	we arrive at the univariate polynomial in few steps
	and we directly use simplifications arising from intersecting and parallel axes.

Beyond its performance benefits, IK-Geo is open-source and convenient to use.
Implementing the subproblem decomposition algorithms takes few lines of code,
	so the code is easy to understand, modify, port, and debug.
The IK solutions are not based on symbolic simplification exploration, so there is no setup or code generation step.
In fact, the same code can apply to any robot in the same kinematic family,
	even if the exact link lengths or joint offsets are different.

\setlength\tabcolsep{3pt} % narrower vertical spacing (default value: 6pt)
\begin{table}[t]
    \centering
    \caption{Subproblem Formulations}
    \begin{threeparttable}
    \begin{tabular}{@{} r @{\ } l  l}
    \toprule
     & Subproblem &  Equation\\
     \midrule
     1:& Circle and Point & \(\min \norm{\rot(k,\theta)p_1 - p_2}\)\\\addlinespace[0.5em]
     2:& Two Circles & \(\min \norm{\rot(k_1,\theta_1) p_1-\rot(k_2,\theta_2)p_2}\)\\\addlinespace[0.5em]
     3:& Circle and Sphere & \(\min \abs{\norm{\rot(k,\theta)p_1-p_2}-d}\)\\\addlinespace[0.5em]
     4:& Circle and Plane & \(\min \abs{h\tr\rot(k,\theta)p-d}\)\\\addlinespace[0.5em]
     5:& Three Circles\tnote{*} & \(p_0+\rot(k_1,\theta_1) p_1 = \rot(k_2,\theta_2)(p_2+\rot(k_3,\theta_3)p_3)\)\\\addlinespace[0.5em]
     6:& Four Circles\tnote{*} & \(\begin{cases}
    h_1\tr \rot(k_1, \theta_1)p_1 + h_2\tr \rot(k_2, \theta_2)p_2 = d_1\\\addlinespace[0.5em]
    h_3\tr \rot(k_3, \theta_1)p_3 + h_4\tr \rot(k_4, \theta_2)p_4 = d_2
\end{cases}\)\\
     \bottomrule
    \end{tabular}
    \label{tab:subproblems}
    \begin{tablenotes}
        \item[*] Must return continuous approximate solutions.
    \end{tablenotes}
    \end{threeparttable}
\end{table}
\setlength\tabcolsep{6pt} % return to default

To solve the IK problem for any 6R arm, we identify and solve six canonical geometric subproblems, shown in Table~\ref{tab:subproblems}, while also optimizing computational efficiency and precision.
Subproblems 1--3 are the original Paden-Kahan subproblems~\cite{Paden:M86/5, Kahan83}.
We are careful to pose the subproblems so that they are robust to singularities through a continuity requirement,
	and to our knowledge we are the first to solve these subproblems in this way.
We pose the subproblems to return an approximate solution for any branch without an exact solution, and the mapping between the subproblem parameters and the returned solutions must be continuous. For Subproblems~1--4, we make this requirement stronger and require least-squares solutions.
Subproblems must also identify and solve cases where there is a continuum of solutions.

In the subproblem formulations,
    \(\theta\) is an unknown angle,
    \(p\) is an \(\rr3\) vector,
    \(h\) is a unit \(\rr3\) vector,
    and \(d\) is a scalar.
    (\(d\) must be nonnegative in Subproblem~3.)
Note that $\rot(k,\theta)$ denotes the rotation matrix about a unit vector $k\in\rr 3$ over an angle $\theta$.
For some fixed \(p, k\) and parameter \(\theta\), \(\rot(k,\theta)p\) sweeps out the circle on the edge of a right circular cone with axis \(k\) and generator \(p\).
We also consider generalizations of Subproblem~2 where the circle axes do not intersect.
As long as there is no continuum of solutions,
Subproblem~1 has
    one exact solution or
    one least-squares solution,
Subproblems~2, 3, and 4 have
    up to two exact solutions or
    one least-squares solutions (Subproblem 2 has up to two least-squares solution if \(\norm {p_1} \neq \norm {p_2}\)),
and Subproblems~5 and 6 have
    up to four solutions.

Our subproblem solution method is based on geometry and linear algebra and does not rely on symbolic manipulation.
	By using ATAN2 rather than \(\sin^{-1}\), \(\cos^{-1}\), or the tangent half angle substitution,
		we improve precision and prevent edge cases where the solution method fails.
	Subproblems~2 and 3 are solved using Subproblem~4.
	Subproblems~5 and 6 are reduced to solving the intersection of two ellipses,
		which may be solved by finding roots of quartic polynomials,
		and approximate solutions which are continuous with respect to the input are found by taking the real part of complex pairs of solutions.
To better identify singularities and to increase computational efficiency, 
	we explain conditions under which the subproblems may be decomposed further.

Beyond 6R IK,
the subproblem decomposition method may be used for
	forward and inverse kinematics for parallel robots,
	IK for robots with prismatic joints,
	IK for robots with a different number of DOF such 7-DOF redundant manipulators~\cite{elias2023redundancy},
	and other geometric problems such as in computer graphics.

The remaining sections of this paper are as follows. %
We provide an overview of the history of inverse kinematics and the subproblem decomposition method in Section~\ref{sec:related_work},
and then explain our IK approach in Section~\ref{sec:approach}.
In Sections~\ref{sec:closed_form_IK} and \ref{sec:search_based_IK}, we provide closed-form and search-based IK solutions,
and in Section~\ref{sec:evaluation} we demonstrate significant performance speedups compared to alternatives like IKFast.
In Section~\ref{sec:polynomial} we explain our method of IK by finding zeros of the  polynomial in the tangent half angle.
We conclude and discuss ideas for improving the algorithms even further in Section~\ref{sec:conclusion}.
The six subproblems are solved in Appendix~\ref{sec:app_subproblem_solns}, and 
numerical examples for the polynomial approach are shown in Appendix~\ref{sec:app_num_examples}.

The IK solutions, subproblem solutions, and timing tests are coded in MATLAB and Rust.
The polynomial method is implemented in Mathematica and Maple.
All code is opens source and available on a publicly accessible repository\footnote{\url{https://github.com/rpiRobotics/ik-geo}}.

\section{Related Work} \label{sec:related_work}
This work is a continuation of a long line of work in analytical and semi-analytical (i.e., reduced-dimension search) solutions to the inverse kinematics problem. The seminal thesis by Pieper in 1969 \cite{pieper1969kinematics} presented analytical solutions for many special robot cases, and for the general 6R robot case, he showed the polynomial in the tangent half angle is at most degree 524,288. Since then, work on the IK problem by other authors has tended to focus on two areas: The general 6R robot IK problem and simplified kinematics IK problems.

To solve the general 6R robot IK problem, most authors find a 16th order polynomial whose real roots correspond to the (up to) 16 solutions for such an arm. Although the general 6-DOF IK problem has been solved, solutions often require computer-assisted algebra or the tangent half angle substitution, so there is still a need for more robust, faster, and more easily understood algorithms. Much of this work was published in the early 1990s. Raghavan and Roth \cite{raghavan1990kinematic, raghavan1993inverse} were one of the first to devise a procedure to construct a 16th order polynomial in the tangent half angle of one of the joints. Many others quickly followed with simpler and more efficient procedures.
\cite{lee1991complete} found a 16th order polynomial in a different manner, \cite{kohli1993inverse} reduced general 6R IK to an eigenvalue problem or generalized eigenvalue problem, and \cite{manocha1994efficient} reduced the problem to a \(12\times12\) quadratic eigenvalue problem.
More recently in 2007, \cite{husty2007new} solved the general 6R IK problem using dual quaternions and 7D projective geometry. They reduced the IK problem to the intersection of eight hyperplanes and a six-dimensional quadric, which is solved by finding 48-degree polynomial and factoring out a 16th order polynomial.
A 2D search-based method is discussed in \cite{fundamentalsBook} (they call it the bivariate-equation approach) where two joint angles are searched over, and the tangent half angle substitution is not used.
In \cite{rudny2014solving}, the 2D search method presented by \cite{fundamentalsBook} was extended by efficiently finding all solutions using Bernstein elimination.

For certain robot geometries, there are fewer than 16 solutions, and the order of the polynomial in the tangent half angle decreases \cite{mavroidis1994structural}. In some cases, the solution may be simplified down to solving quartic or even quadratic polynomials. While using the general 6R methods for simplified robot cases is viable, the solutions tend to be much easier when using methods specific to those cases. Pieper \cite{pieper1969kinematics} focused much of his work on the cases of three intersecting axes because the rotation and position parts of the IK problem become decoupled. The three parallel axes case is similar, as the three axes still intersect but at a point at infinity. Search may also be used to solve such simpler robots, and in \cite{xiao2021effective, trinh2015geometrical}, 1D search is used to solve IK for some simple manipulators with many intersecting or parallel axes

Paden \cite{Paden:M86/5} showed that for some of these special 6R robots, the inverse kinematics problem may be decomposed into a series of smaller canonical subproblems with closed-form solutions.  The influential robotics textbook \cite{MurrayLiSastry} called these subproblems the Paden--Kahan Subproblems 1, 2, and 3.  However, Subproblems~1--3 alone cannot be used to solve all 6R robots with three parallel or intersecting axes. We identify Subproblems~5 and 6 as the canonical subproblems solved in the case of three intersecting or parallel axes, respectively.
Subproblem~5 has been identified by many names including the position problem \cite{fundamentalsBook}, the third order subproblem \cite{wang2018general_frame}, the three-joint subproblem \cite{song2022general}, and the wrist center positioning problem \cite{zsombor20093r}.
We also identify Subproblem~4 as a canonical subproblem, as it is used when there are two parallel axes in a robot or in solving a spherical wrist.

Various authors have contributed solutions to the original three Paden-Kahan Subproblems as well as Subproblems~4--6, either by explicitly identifying them as subproblem solutions or by addressing them while solving IK problems.
(Authors tend to solve the exact versions rather than the least-squares or continuous versions of these subproblems.)
In \cite{dimovski2018algorithmic, yue2008extension, chen2015improved, an2018generalized, wang2018novel, wang2018general_frame, kong2006solution, xu2019models, chen2001closed, leoro2021new, zhao2008generation, wang2018general_inverse, sariyildiz2011comparison, tan2010solution, tan2011inverse, pardos2021screw, song2022general}, subproblems are solved using screw theory.
In \cite{sariyildiz2011comparison}, subproblem solutions are provided with quaternions, and in \cite{lin2019analytical,  chen2020solution, josuet2016improved, sariyildiz2011comparison} dual quaternions are used.
In \cite{yue2008extension,an2018generalized, wang2018novel, wang2018general_inverse, pardos2021screw}, a new subproblem of two consecutive rotations with an offset is posed, which is a generalization Subproblem~2.
In \cite{leoro2021new, josuet2016improved, pardos2021screw, kong2006solution}, the solution for two consecutive parallel rotations is provided, which is a special case of this generalization.
Subproblem 4 has been solved before as part of the solution to the inverse kinematics of a spherical wrist, such as in \cite{shuster2003generalization}.
Several authors posed new subproblems with three consecutive rotations with constraints on which axes are parallel or intersecting, which are special cases of Subproblem 5 \cite{dimovski2018algorithmic, chen2015improved,  xu2019models, chen2020solution, leoro2021new,  wang2018general_inverse, pardos2021screw, kong2006solution}.
In \cite{fundamentalsBook, zsombor20093r, wang2018general_frame, song2022general}, the fully general case of three consecutive rotations is solved. In \cite{clement2013motoman}, a specific instance of Subproblem~5 was solved for IK of a 7R arm, and the authors noticed that the problem reduces to the intersection of two ellipses. The general 6-DOF methods presented in \cite{husty2007new} simplify when a robot has three intersecting or parallel axes, meaning this method can solve Subproblems~5 and 6, as elaborated on in \cite{pfurner2009explicit}.  In solving Subproblem~5, the simplifying cases of two parallel or intersecting axes means authors also solve Subproblems~2 and 4. In \cite{fundamentalsBook}, the author reduces Subproblem~5 to Subproblem~6, and so inadvertently solves Subproblem~6 as well.

To demonstrate the computational performance and range of robot kinematics that the subproblem decomposition method offers, we will compare performance to IKFast and to a MATLAB implementation of Pieper's method.
The original implementation of IKFast~\cite{diankov_thesis} could find IK solutions for robots with three intersecting axes by decomposing the kinematics into a position and rotation part and solving them separately using symbolic manipulation. More recent updates to the program code also include implementations of the general 6-DOF IK algorithms presented in \cite{lee1991complete, kohli1993inverse, manocha1994efficient}.
MATLAB's Robotics System Toolbox~\cite{matlabRobotics} closed-form inverse kinematics solver, \verb|analyticalInverseKinematics|, is based on the Pieper method \cite{pieper1969kinematics} and is designed to work with any robot with a spherical wrist. The solver generates a MATLAB function which finds all inverse kinematics solutions, which can then be converted to a C++ MEX function with MATLAB Coder.

\section{Subproblem Decomposition Approach to IK}\label{sec:approach}
\begin{figure}[tb]
    \centering
    \includegraphics[scale=0.5, clip]{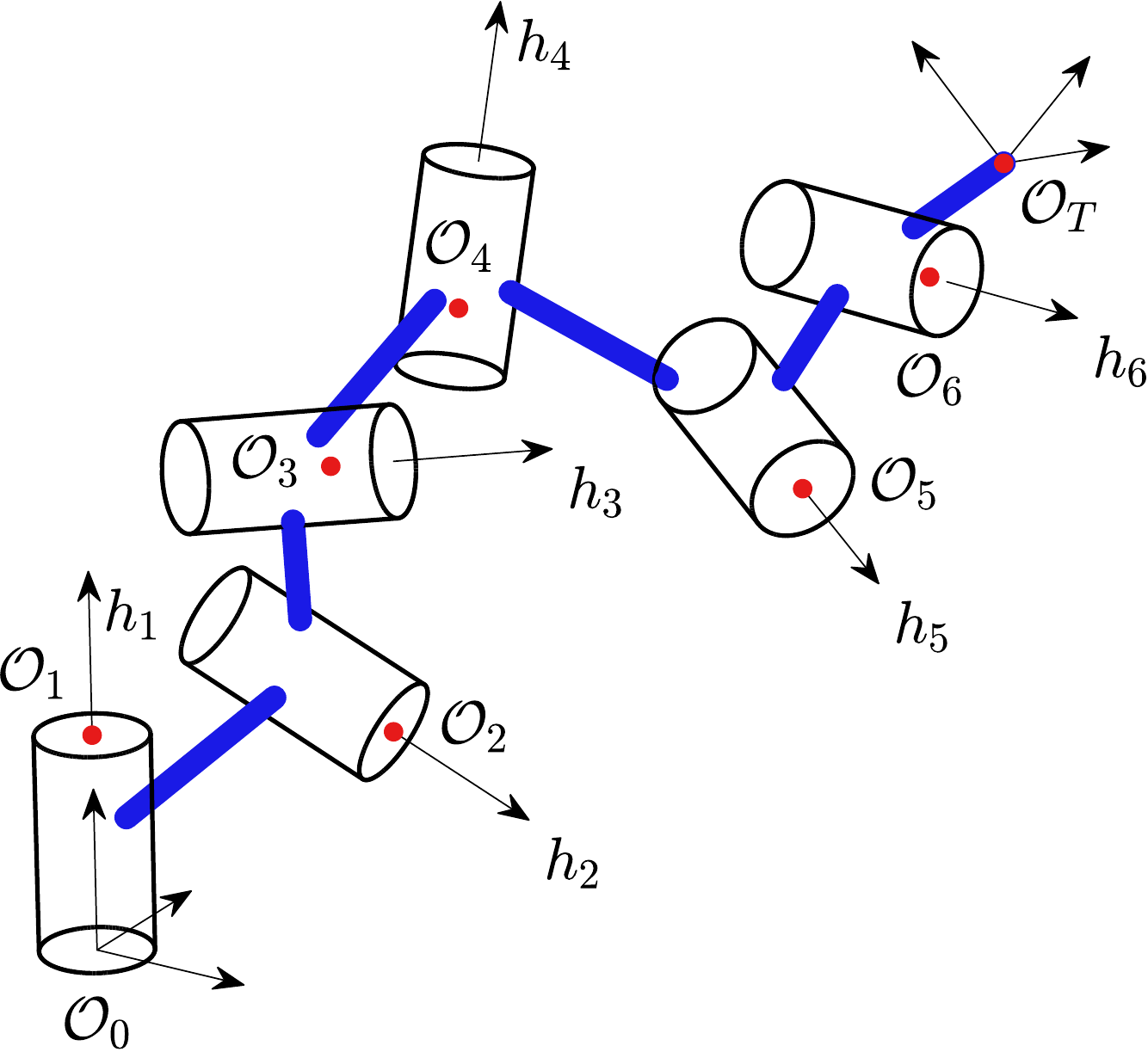}
    \caption{General 6R robot arm with $h_i$ as the $i$th joint axes and ${\mathcal O}_i$ as the origin of frame \(i\), placed anywhere along $h_i$. Inverse kinematics can be solved using Subproblem 5 and a 2D search over two of the joint angles.}
    \label{fig:general6dof}
\end{figure}

Consider a six-DOF articulated robot with all revolute joints as shown in Fig.~\ref{fig:general6dof}. We will use the product of exponentials convention to describe the arm kinematics:
Let frame \(0\) be the world coordinate frame, let frame \(i, i=1,\ldots, 6\) be the coordinate frame representing the orientation of link \(i\) which has equal orientation to frame \(0\) when the robot is in the zero configuration, and let frame \(T\) be the coordinate frame representing the orientation of the end effector tool.
Let \(\mathcal O_0\) be the origin of the world frame, let \(\mathcal O_i\), \(i=1,\ldots, 6\), be the origin of frame \(i\), placed anywhere along the \(i\)th joint axis, and let \(\mathcal O_T\) be the origin of the end effector tool frame.
Define \(p_{i,j}\) (\(p_{i,T}\)),  \(i,j=0,\ldots, 6\), be the $\rr3$ vector from  \(\mathcal O_{i}\) to \(\mathcal O_{j}\) (\(\mathcal O_{T}\)) represented in the \(i\) frame. This means \(p_{i-1,i}\), \(i=1,\ldots,6\), and \(p_{6,T}\) are constant $\rr3$ vectors.

Define \(h_i\) to be the constant unit vector rotation axis for the $i$th joint represented in the \(i\) frame, and let $R_{i-1,i}=\rot(h_i,q_i)$, where $q_i$ is the $i$th joint angle. Also let $R_{i,i-1}=\rot(h_i,-q_i)$, and for $i<j$, $R_{ij}=R_{i,i+1}\cdots R_{j-1,j}$.  Finally, define \(R_{6T}\) to be the constant orientation of the tool frame with respect to the 6 frame. Then, the rotation and position of the tool frame with respect to the world frame are
\begin{subequations}
    \begin{align}
        R_{0T} ={}&R_{01} R_{12} R_{23} R_{34} R_{45} R_{56} R_{6T},\\
        \begin{split}
    p_{0T} ={}& p_{01} + R_{01} p_{12} + R_{02} p_{23}+ R_{03} p_{34}\\
            &+ R_{04} p_{45} + R_{05} p_{56} + R_{06} p_{6T}.  
        \end{split}
    \end{align}
    \label{eq:poe_fwdkin}
\end{subequations}

The inverse kinematics problem for a robot with kinematic parameters \(\left( \{p_{i-1,i}\}_{i=1}^6,\ p_{6T},\ \{h_i\}_{i=0}^6,\ R_{6T} \right)\) means finding all joint angles \(q = \begin{bmatrix}
q_1 & q_2 & \cdots & q_6
\end{bmatrix}\tr\) corresponding to a given end effector rotation and position \((R_{0T},p_{0T})\). In general, one \((R_{0T},p_{0T})\) corresponds to multiple \(q\), which are called the different inverse kinematics branches.

Without loss of generality, assume \(p_{01} = 0\). (Otherwise, subtract \(p_{01}\) from \(p_{0T}\).)  Rewriting \eqref{eq:poe_fwdkin} by moving constants to the left-hand side results in
\begin{subequations} 
    \begin{align}
    R_{06} ={}& R_{0T}R_{6T}\tr = R_{01} R_{12} R_{23} R_{34} R_{45} R_{56},\label{eq:poe_fwdkin_simplified_rot}\\ 
    \begin{split} \label{eq:poe_fwdkin_simplified_pos}
    p_{06} ={}& p_{0T}-R_{06}p_{6T} \\
    ={}& R_{01} p_{12} + R_{02} p_{23}+ R_{03} p_{34}\\
            &+ R_{04} p_{45} + R_{05} p_{56}. 
   \end{split}
    \end{align}
    \label{eq:poe_fwdkin_simplified}
\end{subequations}
This means the IK procedures can be in terms of \((R_{06}, p_{06})\).

In the following inverse kinematics solutions using the subproblem decomposition method, we start with \eqref{eq:poe_fwdkin_simplified} and simplify based on any intersecting axes: if axes \(i-1\) and \(i\) intersect, we can pick \(\mathcal O_{i-1} = \mathcal O_{i}\), and so \(p_{i-1,i} = 0\).
Then, we decompose the equations into subproblems by moving terms and multiplying by rotations, taking the norm of both sides of the equation, or, in the case of parallel axes, projecting both sides of the equation onto a vector by left multiplying by \(h_i\tr\) or right multiplying by \(h_i\). 
This corresponds to solving the equation \(\rot(k,\theta)p_1 = p_2\) (where \(p_1\) and \(p_2\) may be functions of other unknowns) by solving the necessary condition \(\norm {p_1} = \norm {p_2}\) or \(k\tr p_1 = k\tr p_2\). In general, these conditions are not sufficient, and so solutions must be checked if they are extraneous. However, any extraneous solutions earlier in the procedure become apparent in the next subproblems, as they do not return any exact solutions.

In some cases, the entire IK problem can be decomposed into only subproblems, which leads to closed-form solutions. In other cases, the decomposition remains in terms of one or two joint angles, and so a search method can be used. We also show in Section~\ref{sec:polynomial} that rather than search, we can keep the same decomposition but use the tangent half angle substitution to convert the decomposition to a system of polynomials.

When a subproblem has multiple solutions, the solution for \(q\) branches, meaning the procedures find all inverse kinematics solutions. (If desired, just one of the solutions can be used to speed up computation time.) If a subproblem has a continuum of solutions, such as when two rotation axes are collinear, then the robot has an internal singularity for that branch. This method is robust to such singularities as the arbitrary joint angles can be detected and resolved before continuing the procedure. If a subproblem has no exact solutions, then that branch fails to find a solution; however, using the least-squares solution for Subproblems~1--4 and the similar continuous approximate solutions for Subproblems~5 and 6 are still useful as described in Section~\ref{sec:LS_IK}.

The following solutions are valid not just for a single robot, but for a family of robots with certain conditions on their kinematics parameters based on consecutive intersecting or parallel joints. Therefore, a single code implementation can perform inverse kinematics for multiple robots in the same kinematic family but with different dimensions.

In the solutions below we demonstrate specific locations for where interesting or parallel axes occur. However, similar equations can be easily derived for intersecting or parallel axes occurring elsewhere in the chain.
\begin{figure}
\centering
\subfloat[]{\includegraphics[scale=0.5, clip, trim={0.2in 0.2in 0.35in 0.2in}]{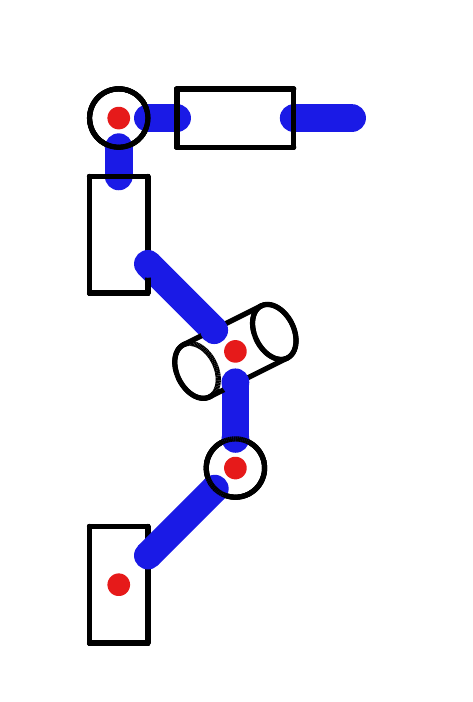}}\quad%
\subfloat[]{\includegraphics[scale=0.5, clip, trim={0.2in 0.2in 0.35in 0.2in}]{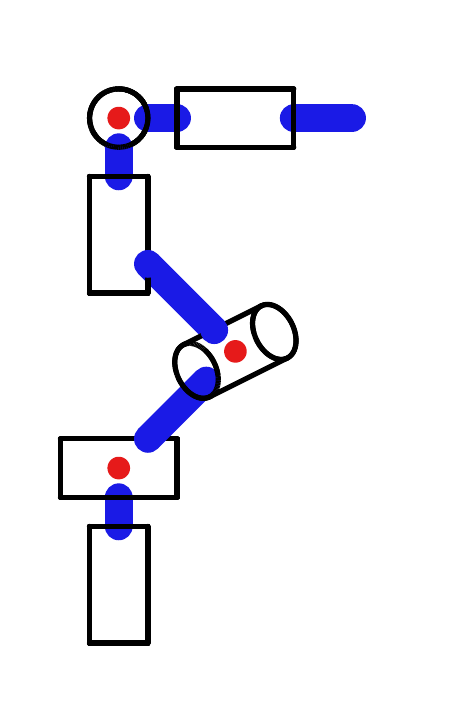}}\quad%
\subfloat[]{\includegraphics[scale=0.5, clip, trim={0.2in 0.2in 0.35in 0.2in}]{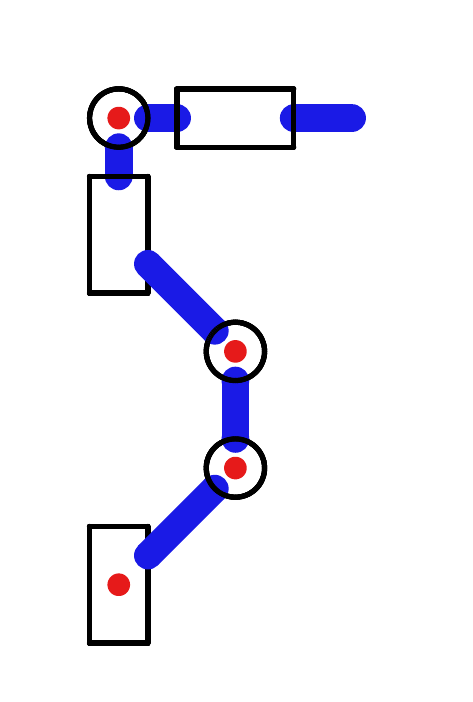}}\\%
\subfloat[]{\includegraphics[scale=0.5, clip, trim={0.2in 0.2in 0.35in 0.2in}]{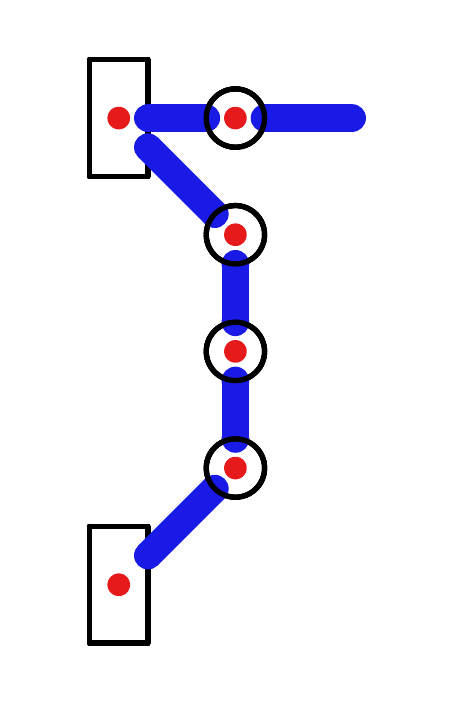}}\quad%
\subfloat[]{\includegraphics[scale=0.5, clip, trim={0.2in 0.2in 0.35in 0.2in}]{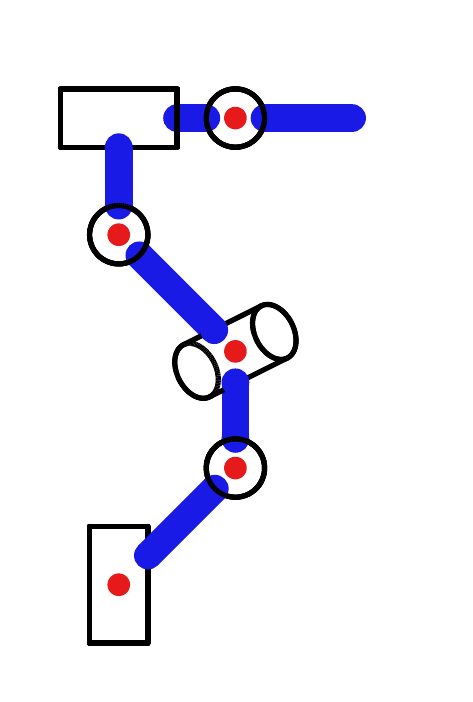}}\quad%
\subfloat[]{\includegraphics[scale=0.5, clip, trim={0.2in 0.2in 0.35in 0.2in}]{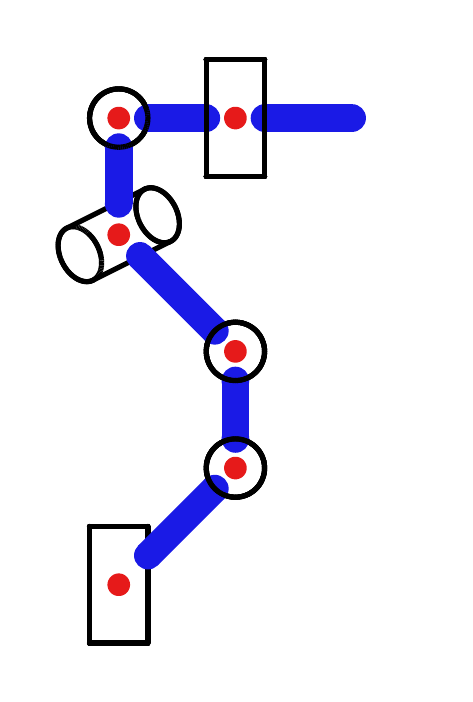}}\quad%
\subfloat[]{\includegraphics[scale=0.5, clip, trim={0.2in 0.2in 0.35in 0.2in}]{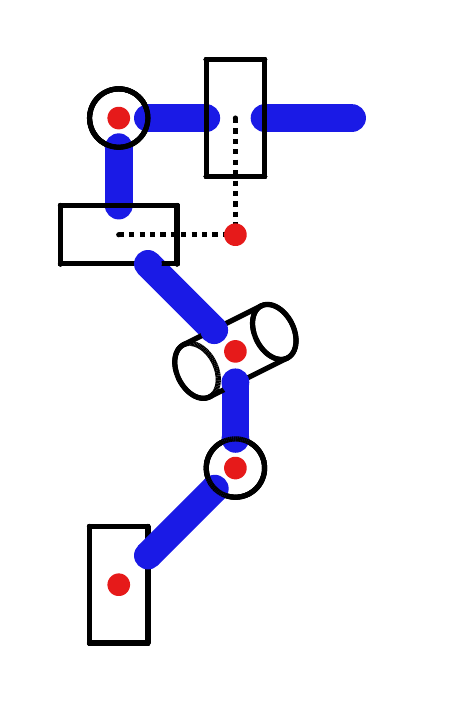}}
\caption{
Some cases for simplified robot inverse kinematics.
(a)~Spherical wrist.
(b)~Spherical wrist and two intersecting axes.
(c)~Spherical wrist and two parallel axes.
(d)~Three parallel axes.
(e)~Two intersecting axes.
(f)~Two parallel axes.
(g)~Two nonconsecutive intersecting axes.
}
\label{fig:robots}
\end{figure}
\section{Closed-Form Inverse Kinematics}\label{sec:closed_form_IK}
If a 6R robot has three consecutive intersecting or parallel joint axes, then the inverse kinematics can be solved using Subproblems 5 or 6, respectively. If a robot has certain other pairs of intersecting or parallel axes, then the solution can sometimes be written in terms of just Subproblems~1--4. This is in accordance with the special simplified cases of Subproblems~5 and 6. 
\subsection{Three Intersecting Joint Axes}\label{sec:three_intersecting}

When a 6R robot has three intersecting joint axes, the inverse kinematics problem can be decoupled into position and rotation parts. The position part can be solved with Subproblem~5, and the rotation part can be solved with Subproblems~1 and 2. Consider the common case of a spherical wrist where the last three joint axes $(h_4,h_5,h_6)$ intersect at a point, shown in Fig.~\ref{fig:robots}(a).
Choose the origins of frames 4, 5, 6 at the intersection so $p_{45}=p_{56}=0$. Then, \eqref{eq:poe_fwdkin_simplified_pos} becomes
\begin{equation}
    -p_{12} + R_{10} p_{06} = R_{12} (p_{23}+R_{23}p_{34}),
    \label{eq:3dofinvkin}
\end{equation}
which only depends on $(q_1,q_2,q_3)$ and can be solved using Subproblem~5.

To solve for the wrist angles  $(q_4,q_5,q_6)$, write \eqref{eq:poe_fwdkin_simplified_rot} as
\begin{equation}
    R_{32}R_{21}R_{10}R_{06} = R_{34}R_{45}R_{56}.
    \label{eq:spherical_wrist_rotation}
\end{equation}
For each $(q_1,q_2,q_3)$, the left-hand side is known, which can be called $R_{36}$.  Project it to $h_6$:
\begin{equation}
    R_{43}R_{36}h_6 = R_{45} h_6,
\end{equation}
and solve for up to two solutions of $(q_4, q_5)$ using Subproblem~2. Then, find \(q_6\) by using Subproblem~1 to solve \eqref{eq:spherical_wrist_rotation}. For a spherical wrist to achieve any orientation, it must have orthogonal consecutive joint axes.

If a robot has two intersecting axes separate from the three intersecting axes, then Subproblem~5 simplifies into Subproblems~2 and 3. Consider a robot with a spherical wrist and the first two axes intersecting as in Fig.~\ref{fig:robots}(b). In this case, $p_{12}=0$. Taking the norm of both sides of \eqref{eq:3dofinvkin}, we have 
\begin{equation}
    \norm{p_{06}} = \norm{p_{23}+R_{23} p_{34}},
\end{equation}
which may be solved using Subproblem~3 for up to two solutions of $q_3$. For each $q_3$, we can use Subproblem~2 to find up to two solutions of $(q_1,q_2)$.

If a robot has two parallel axes separate from the three intersecting axes, then Subproblem~5 simplifies into Subproblems~1, 3 and 4.
Consider a robot with a spherical wrist and the second and third axes parallel ($h_2=h_3$) shown in Fig.~\ref{fig:robots}(c). Rearranging \eqref{eq:3dofinvkin} and projecting onto $h_2$, we have
\begin{equation}
    h_2\tr R_{10} p_{06} = h_2\tr (p_{12}+p_{23}+p_{34}),
\end{equation}
which may be solved using Subproblem~4 for up to two solutions of $q_1$. For each $q_1$, we can use Subproblem~3 to find up to two solutions of $q_3$.  For each of the four possible $(q_1,q_3)$, we can solve for the corresponding $q_2$ using Subproblem~1.

\subsection{Three Parallel Joint Axes} \label{sec:3_parallel}
When a 6R robot has three parallel axes, the inverse kinematics problem can be solved using Subproblems~1, 3, and 6. (Such a robot is the limit of a robot with three intersecting axes where the intersection point moves infinitely far away.) Consider a robot with $h_2=h_3=h_4$ as shown in Fig.~\ref{fig:robots}(d).
Projecting \eqref{eq:poe_fwdkin_simplified_pos} onto $h_2$ results in
\begin{equation}
    h_2\tr R_{10} p_{06}- h_2\tr R_{45}p_{56}=h_2\tr(p_{12}+p_{23}+p_{34}+p_{45}).
    \label{eq:3axescase1}
\end{equation}
From \eqref{eq:poe_fwdkin_simplified_rot}, we have
\begin{equation}
    h_2\tr R_{10} R_{06}h_6 - h_2\tr R_{45} h_6 = 0.
    \label{eq:3axescase2}
\end{equation}
If $p_{56}=0$, we can find up to two solutions of $q_1$ from \eqref{eq:3axescase1} using Subproblem~4. (A similar simplification occurs if a robot has three parallel axes and two other parallel axes.) For each solution of \(q_1\), we can use Subproblem 4 to solve for up to two solutions of $q_5$ from \eqref{eq:3axescase2} for a total of four solutions. If $p_{56} \neq 0$, \eqref{eq:3axescase1} and \eqref{eq:3axescase2} form a system of equations which can be solved using Subproblem 6 for up to four solutions of $(q_1,q_5)$.

By projecting \eqref{eq:poe_fwdkin_simplified_rot} onto \(h_6\) or \(h_2\), we can solve for $q_2+q_3+q_4$ (and hence $R_{14}$) and \(q_6\) using Subproblem~1 to solve
\begin{align}
    R_{14}R_{45}h_6 &= R_{10}R_{06}h_6,\\
    R_{65}R_{54}h_2 & = R_{06}\tr R_{01}h_2.
\end{align}
Find up to two solutions of $q_3$ using Subproblem~3 to solve
\begin{equation}
    \norm{p_{23} + R_{23}p_{34}}
    {=} % shrink horizontal space
    \norm{
        R_{10}p_{06} - p_{12} - R_{14}p_{45} - R_{15}p_{56}
    }
\end{equation}
for a total of eight solutions. Within these eight solutions, we can solve for \(q_2\) using Subproblem 1:
\begin{equation}
    R_{12}(p_{23}+R_{23}p_{34}) = R_{10}p_{06} - p_{12} - R_{14}p_{45} - R_{15}p_{56},
\end{equation}
and we can solve for \(q_4\) using subtraction, wrapping the result to \([-\pi, \pi]\) if desired.

\subsection{Continuous and Least-Squares IK Solutions}\label{sec:LS_IK}
Subproblems~1--4 return least-squares solutions, and Subproblems~5 and 6 also have solution continuity by returning the real part of any complex solutions.
This means the IK procedure returns solutions for \(q\) even for branches which may not have an exact solution. If a desired end effector pose is not feasible but near the robot workspace, the IK procedure will return a close solution. The desired pose can be exactly on the boundary of the workspace (i.e. at a singularity), and the IK procedure will correctly find the robot joint angles even in the presence of numerical inaccuracies which may slightly perturb the desired pose outside workspace.

For certain robots with the task frame placed at the wrist, the least-squares solutions from Subproblems~1--4 also solve the global least-squares problem in the usual Euclidean sense, solving the minimization
\begin{equation} \label{eq:IK_LS_minimization}
    \min_q \norm{p_{0T}(q) - p_{0T}^{des}} \text{ s.t. } R_{0T}(q) = R_{0T}^{des},
\end{equation}
where \((R_{0T}^{des}, p_{0T}^{des})\) is the desired end effector rotation and position. 
One robot configuration that achieves \eqref{eq:IK_LS_minimization} is the  common case of industrial robots similar to the ABB IRB 6640~\cite{ABB_IRB_6640}: a 6R robot with a spherical wrist (\(p_{45}=p_{56}=0\)) and two parallel axes (\(h_2 = h_3\)), as well as \(h_2\tr (p_{12} + p_{23} + p_{34}) = 0\), \(h_1\tr h_2 = 0\), and \(p_{6T}= 0\). We also need \(h_4 \tr h_5 = h_5 \tr h_6 = 0\) so the spherical wrist can achieve any orientation. Under these conditions, we are minimizing distances in a cylindrical coordinate system.
For some robots, such as when we have all the constraints above except \(h_1\tr h_2 \neq 0\), \eqref{eq:IK_LS_minimization} is solved in only certain regions of the workspace.

\section{Search-Based Inverse Kinematics}\label{sec:search_based_IK}
If at least one pair of consecutive joints are intersecting or parallel then the problem can be reduced to a 1D search over a circle using Subproblems 5 or 6, respectively. If a pair of nonconsecutive joint \(k, k+2\) always intersect, then problem can also be solved with a 1D search. If multiple pairs of joints are parallel or intersecting, then sometimes a 1D search may be performed using only Subproblems~1--4. In the fully general case where no consecutive joints are parallel or intersecting, then the problem reduces to a 2D search over a torus using Subproblem 5. 

\subsection{Two Intersecting Joint Axes (1D Search)}\label{sec:2_intersecting}
\begin{figure}
    \centering
    \includegraphics[width=\linewidth]{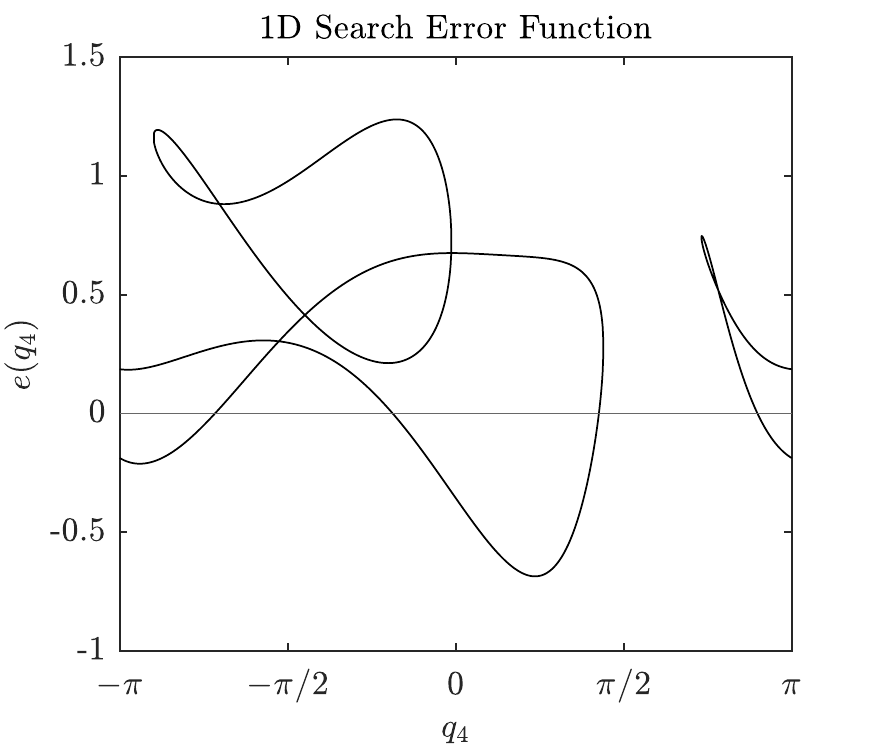}
    \caption{Error function for a robot with axes 5 and 6 intersecting. For this end effector pose there are four IK solutions corresponding to the four zeros of the error function.  There are four branches in this error function corresponding to the four solutions of Subproblem~5. For a given input, the error function has anywhere from zero values (e.g., \(q_4 = \pi/2\)) to four values (e.g., \(q_4 = -\pi/2\)).}
    \label{fig:search_err}
\end{figure}
A 6R robot with two intersecting axes can be solved by searching over one joint angle.
If $p_{56}=0$ as in Fig.~\ref{fig:robots}(e), then \eqref{eq:poe_fwdkin_simplified_pos} becomes
\begin{equation}\label{eq:sp5_two_intersecting}
    -p_{12} + R_{10}p_{06} = R_{12}\left(
        p_{23} + R_{23} \left( p_{34} + R_{34}p_{45}\right)
    \right).
\end{equation}
In this case, for each $q_4$, we can find up to four solutions of $(q_1,q_2,q_3)$ using Subproblem 5. 
The error measures the solvability of
\begin{equation}
    R_{45}R_{56} = R_{04}\tr R_{06}.
\end{equation}
By projecting onto \(h_5\) and \(h_6\), we get the error
\begin{equation} \label{eq:err_2_intersecting}
    e(q_4) = h_5\tr R_{04}\tr R_{06}h_6 - h_5 \tr h_6.
\end{equation}
Perform a 1D search to find the zeros of \eqref{eq:err_2_intersecting}, as shown in Fig.~\ref{fig:search_err}, and use Subproblem~1 to find \(q_5\) and \(q_6\) according to
\begin{align}
    R_{45} h_6 &= R_{04}\tr R_{06} h_6,\\
    R_{65} h_5 &= R_{06}\tr R_{04} h_5.
\end{align}
Geometrically, 
    we can interpret this search method as replacing the original robot with a modified robot with joint 4 fixed and an extra joint added to the intersection of joints 5 and 6. This creates a new modified robot with a spherical wrist and closed-form IK. If the IK solution for this modified robot has the new joint at zero angle, then this is a solution to the IK problem for the original robot.
    
\subsection{Two Parallel Joint Axes (1D Search)}\label{sec:2_parallel}
Since a robot with two parallel axes is the limit of a robot with two intersecting axes where the point of intersection moves to a point infinitely far away, such a robot can also be solved with a 1D search.
If \(h_2 = h_3\) as in Fig.~\ref{fig:robots}(f), then by substituting \(R_{56} = R_{06}R_{65}\) and left multiplying by \(h_2 \tr R_{10}\), \eqref{eq:poe_fwdkin_simplified_pos} becomes
\begin{multline}
     h_2 \tr R_{10}  R_{06} R_{65} p_{56}
    + h_2 \tr R_{34} p_{45}\\
    = h_2 \tr ( R_{10} p_{06} -p_{12} - p_{23} - p_{34}) .
\end{multline}
Projecting \eqref{eq:poe_fwdkin_simplified_rot} onto \(h_2\) and \(h_5\) yields
\begin{equation}
    h_2\tr R_{10} R_{06} R_{65} h_5 - h_2\tr R_{34} h_5 = 0.
\end{equation}
Search over \(q_1\) and use Subproblem~6 to find \((q_4, q_6)\).
Next, find \(q_2 + q_3\) and therefore \(R_{13}\) with Subproblem~1 to solve
\begin{equation}
    R_{13} R_{34}h_5 = R_{10} R_{06} R_{65} h_5.
\end{equation}
The error measures the solvability of
\begin{align}
    \begin{split}
    R_{12} p_{23}
    ={}& R_{10} p_{06}
    - p_{12}
    - R_{13}p_{34}\\
    &{}- R_{14}p_{45}
    - R_{10} R_{06}R_{65} p_{56}.
    \end{split}
\end{align}
We search for \(q_1\) to find the zeros of the error
\begin{align}
    \begin{split}
    e(q_1) =\lVert& R_{10} p_{06}
    - p_{12}
    - R_{13}p_{34}\\
    &- R_{14}p_{45}
    - R_{10} R_{06}R_{65} p_{56}\rVert
    - \norm{p_{23}}.
    \end{split}
\end{align}
Then solve for \(q_2\) and \(q_5\) with Subproblem~1, solving
\begin{equation}
    h_2\tr R_{34} R_{45} = h_2\tr R_{10} R_{06} R_{65},
\end{equation}
\begin{equation}
    R_{12}p_{23} = R_{10} p_{06} - p_{12} - R_{13}p_{34} - R_{14}p_{45} - R_{15}p_{56},
\end{equation}
and calculate \(q_3\) with subtraction, wrapping the result to \([-\pi, \pi]\) if desired.

The geometric interpretation of this search method is similar to the two intersecting case. We replace the original robot with new modified robot with joint 1 fixed and a new prismatic joint added between joints 2 and 3. We solve IK for this modified robot in closed form, and if the new joint has zero displacement, then this is a solution to the IK problem for the original robot.

\subsection{Two Nonconsecutive Intersecting Axes (1D Search)} \label{sec:2_nonconsecutive_intersecting}
If a 6R robot has nonconsecutive axes \(k\) and \(k+2\) always intersecting, then IK can be solved by searching over \(q_{k+1}\) in almost the same way as in Section~\ref{sec:2_intersecting}.
If axes 4 and 6 always intersect as in Fig.~\ref{fig:robots}(f) then for a given \(q_5\) we can find the point of intersection \(O_4' = O_6'\) by shifting \(\mathcal O_4\) along axis 4 and \(\mathcal O_6\) along the rotated axis 6. This gives us two new vectors \(p_{04}'\) and \(p_{34}'\). We find \((q_1, q_2, q_3)\) by using Subproblem~5 to solve
\begin{equation}
    -p_{12} + R_{10}p_{04}' = R_{12}(p_{23} + R_{23} p_{34}').
\end{equation}
The error measures the solvability of 
\begin{equation}
    R_{03} R_{34} R_{45} R_{56}  = R_{06},
\end{equation}
and projecting onto \(h_4\) and \(h_6\) yields
\begin{equation}
    e(q_5) = h_4\tr R_{03}\tr R_{06} h_6 - h_4\tr R_{45} h_6.
\end{equation}
Search over \(q_5\) to find the zeros of this error, then use Subproblem~1 to solve for \(q_4\) and \(q_6\):
\begin{equation}
    R_{34}R_{45}h_6 = R_{03}\tr R_{06} h_6,
\end{equation}
\begin{equation}
    R_{65}R_{54}h_4 = R_{06}\tr R_{03} h_4.
\end{equation}
For some special choices of \(q_5\), axes 4 and 6 are parallel rather than intersecting, and the remaining joint angles must be found with Subproblem~6 rather than Subproblem~5.

\subsection{General Case (2D Search)}\label{sec:gen_6R}
IK for a general 6R robot with no intersecting or parallel axes can be solved by searching over any two joint angles.
Given $(q_1,q_2)$, we can solve for up to four solutions of $(q_3,q_4,q_5)$ using Subproblem~5:
\begin{multline}
    - p_{34} + R_{32}( R_{21}(R_{10}p_{06} - p_{12})-p_{23} )\\
    = R_{34}(p_{45} + R_{45} p_{56}).
    \label{eq:general6R}
\end{multline}
With each $(q_1,q_2,q_3,q_4,q_5)$, compute the vector error
\begin{equation}\label{eq:general6Rerror}
    e(q_1, q_2) =R_{01}R_{12}R_{23}R_{34}R_{45}h_6 - R_{06}h_6.
\end{equation}
Search through $(q_1,q_2)$ to find the zeros of $e$.  For each zero, use Subproblem~1 to solve
\begin{equation}
    R_{56}p = R_{05}\tr R_{06} p
\end{equation}
for $q_6$, where $p$ is any vector not collinear with $h_6$. 

The geometric interpretation for this search method is that we replace the original robot with a new modified robot with fixed joints 1 and 2, and two new joints added to the origin of joint 6. We solve this new robot in closed form, and if the new joints have zero angle, this is a solution to the IK problem for the original robot.

\subsection{Search Implementation}

For both 1D and 2D searches, the error function has multiple branches and may be undefined for some inputs, as shown in Fig.~\ref{fig:search_err}. 
Points with zero slope correspond to robot singularities, and points of infinite slope, which occur at the domain boundary for a given branch, correspond to singularities of the \textit{modified} robot from the geometric interpretation. The slope of the 1D search error and the Jacobian of the 2D search error may be used to measure closeness to singularities.

Performing a 1D search is computationally efficient because the scalar error \(e(q_i)\) crosses zero. One method is to sample the entire search space to find intervals where the error function switches sign. As long as the error function is defined and continuous throughout, zeros of the error function will be in this bracket. We can use a bracketing method such as the false position method, as summarized in \cite{abbasbandy2008new}, to efficiently find the zero. Zeros will be missed if the error function touches but does not cross zero, but this can be fixed by performing a minimization or maximization.

To perform a 2D search, we can find the zeros of \(\norm{e(q_i, q_j)}\). The zeros are also local minima since this is a norm. We first sample the entire search space in a uniform grid and then perform numeric minimization starting with the smallest-valued sample. Once a zero is found, any neighboring samples smaller than some threshold are removed from consideration, and another minimization can be performed starting at the next smallest-valued sample until all samples below some threshold are processed.

A caveat of these particular search implementations is that certain solutions may only be found with a sufficiently small sampling interval.
If we only need to return a single solution for \(q\) and have an initial guess (e.g., to find a solution closest to a previous pose), finding the zero closest to the initial guess does not require sampling the whole search space. This dramatically speeds up the computation time.
Computation time can also be reduced by only searching within joint limits, and it may be useful to perform the search over joints with the smallest ranges of motion.

\section{Evaluation}\label{sec:evaluation}

\begin{table}
    \centering
    \caption{Subproblem Solution Runtimes}
    \begin{tabular}{r @{\ } l  *{5}{r} }
    \toprule
     & & \multicolumn{3}{c}{Mean Runtime (ns)}\\
     \cmidrule{3-5}
        &Subproblem 
            & {m-file}
            & {MEX}
            & {Rust}
            \\ \midrule
     1:& Circle and Point    &  13\,517 &     36 &     59 \\
     2:& Two Circles         &  27\,262 &    312 &    242 \\
     3:& Circle and Sphere   &  18\,034 &    134 &    111 \\
     4:& Circle and Plane    &  17\,669 &    105 &    115 \\
     5:& Three Circles       & 105\,288 & 2\,553 & 1\,027 \\
     6:& Four Circles        & 116\,750 & 3\,110 & 1\,436 \\
     \bottomrule
    \end{tabular}
    \label{tab:subproblem_runtimes}
\end{table}

\begin{table*} 
    \centering
    \caption{Robot Inverse Kinematics Timing}
    \begin{threeparttable}
    \begin{tabular}{ l *{7}{r} r}
    \toprule
                                        & \multicolumn{3}{c}{Mean Random Robot Kinematics Runtime (\textmu s)}     & \multicolumn{4}{c}{Mean Hardcoded Robot Kinematics Runtime (\textmu s)} \\
                                        \cmidrule(r){2-4}\cmidrule(l){5-8}
     Robot Kinematic Family                         & m-file (ours)    & MEX (ours)       & Rust (ours)       & MEX (ours)         & Rust (ours)       & IKFast        & MRT MEX      \\ \midrule
     Spherical wrist                   &       489.893 &      5.158 &                 2.296 &      4.527  &                  2.425 &      111.445  &       10.583 \\
     \quad and two intersecting axes   &       587.320 &      3.423 &                 2.062 &      4.991  &                  3.688 & N/A\tnote{b}  & N/A\tnote{a} \\
     \quad and two parallel axes       &       745.547 &      3.349 &                 2.193 &      2.991  &                  3.183 &        4.417  &       13.318 \\
     Three parallel axes               &       391.796 &      5.762 &                 3.198 &      5.345  &                  4.339 & N/A\tnote{c}  & N/A\tnote{a} \\
     \quad and two intersecting axes   &       376.290 &      2.981 &                 2.232 &      2.846  &                  3.089 &      121.714  & N/A\tnote{a} \\
     Two intersecting axes (1D search) &      35\,592.452\tnote{**} &               659.130 &     339.214 &               199.194  &       114.552 & N/A\tnote{a}  & N/A\tnote{a} \\
     Two parallel axes     (1D search) &      58\,483.194\tnote{**} &               954.959 &     476.725 &            1\,021.655  &       439.445 & N/A\tnote{a}  & N/A\tnote{a} \\
     General 6R (2D search)            &   1\,490\,518.240\tnote{*} & 27\,904.571\tnote{**} & 18\,808.249 & 20\,741.535\tnote{**}  &   16\,026.010 & N/A\tnote{a}  & N/A\tnote{a} \\
     \bottomrule
    \end{tabular}
    \begin{tablenotes}[para]
        \item[*] 100 trials.
        \item[**] 1000 trials.
        \item[a] Did not generate code.
        \item[b] Did not compile.
        \item[c] Incorrect IK.
    \end{tablenotes}
    \end{threeparttable}
    \label{tab:IK_timing}
\end{table*}

Subproblem and IK solutions are implemented in MATLAB and Rust. Additionally, MATLAB code has been compiled to C++ MEX code using MATLAB coder, which can be run from MATLAB.
IK solutions are implemented for eight robot kinematic families. Five solutions are closed-form and return all solutions, two use 1D search and return multiple solutions with 200 initial samples, and one uses 2D search and returns multiple solutions with a grid of 100 samples in each coordinate for a total of 10\,000 initial samples.

For each subproblem or IK solution, testing code generates a file with 10\,000 random test cases (or fewer for certain IK solutions), each with at least one exact solution. After loading the test cases into memory, the execution times and the returned solutions for all test cases are saved.
Robot IK runtimes are measured in two ways. One way was is to randomize both the robot kinematics parameters and the end effector pose. Another way is to hardcode the kinematic parameters into the program and only randomize the end effector pose.
The eight hardcoded robot examples include five robots from Table~\ref{tab:robot_examples} as well as three fictitious robots.

Timing comparisons were also made to two other publicly available analytical IK solvers: IKFast and the MATLAB Robotics Toolbox solver.
IKFast generated valid code for three types of robots but failed in the other five cases.
Generating and compiling code for a robot with three parallel axes and two intersecting axes (UR5) took an exceptionally long time: 55 min and 10 min, respectively.
The MATLAB solver could only find solutions for two robots. The \verb|generateIKFunction| function crashed when generating code for a robot with a spherical wrist and two intersecting axes (KUKA R800 with fixed \(q_3\)).

Testing was performed with MATLAB R2023a running on Windows 10 and \verb|rustc| 1.70.0 running on Windows Subsystem for Linux with Ubuntu 18.04. The computer had an Intel Core i7-3770K CPU at 3.50 GHz and 16 GB memory.

Subproblem timing results are shown in Table~\ref{tab:subproblem_runtimes}.
Subproblems~1--4 ran about an order of magnitude faster than Subproblems 5 and 6, suggesting it is advantageous to use IK algorithms that only use Subproblems~1--4. 
While subproblem solutions using MEX and Rust ran in under 400~ns for Subproblems~1--4 and under 4000~ns for Subproblems 5 and 6, m-file runtimes were about two orders of magnitude slower.

IK timing results are shown in Table~\ref{tab:IK_timing}.
Closed-form IK solutions using the subproblem decomposition method ran in under 6~\textmu{}s for all MEX and Rust implementation, with many solutions running in under 3~\textmu{}s.
Search-based IK was slower than closed-from IK by a factor roughly equal to the number of initial search samples. 
Although m-file implementations ran slower by about two orders of magnitude, they were still fast enough for real-time control.
Hardcoding the kinematics often sped up computation time for MEX code. 
This was most apparent for the general 6R IK since the hardcoded example was the RRC arm with fixed \(q_6\), meaning Subproblem~5 was solved using a quadratic rather than a quartic. Unexpectedly, Rust timing was usually slower for hardcoded kinematics. This is worth investigating as having constants and multiplications by zero should lead to faster compiled code.

MEX and Rust implementations of subproblem decomposition were faster and solved more types of robots than IKFast and Matlab Robotics Toolbox. For example, IK for a robot with three parallel axes and two intersecting axes (UR5) using subproblem decomposition was more than 40 times faster than IKFast.

\section{Connection to Polynomial Method} \label{sec:polynomial}
Rather than using numerical search methods to find the zeros of the error functions for 6R robots without three intersecting or parallel axes, we can instead use symbolic algebra methods to convert the subproblem and error formulations into a polynomial in the tangent half angle of one of the joints. This guarantee finding all solutions to \eqref{eq:poe_fwdkin}, including complex solutions, to arbitrary precision. This method also allows us to detect internal and boundary singularities more easily. The method applies to other search-based solutions which use subproblem decomposition, including parallel robots or cases where the task is not just end effector orientation and position as for 7-DOF robots.

To find the polynomial, we first rewrite the subproblem decomposition and error function using the tangent half angle substitution, which leads to three or four equations in the same number of unknowns. By expanding and clearing the denominators, we obtain a system of multivariate polynomials. The resultant univariate polynomial is found by eliminating all but one variable, and care is taken to factor out extraneous factors which may occur (such as \(x^2+1\)). After finding the zeros of this polynomial, subproblems can then be used to find the remaining joint angles in closed form. In the general 6R case, the zeros of one univariate polynomial must be plugged into an intermediate system of two variables to find solutions to two joint angles before the remaining angles are found in closed form. To improve computational efficiency, we can use rational approximations for the robot kinematics and end effector pose.

Previous literature has found polynomial solutions for the general 6R problem, but the methods presented here have important differences. These methods directly exploit intersecting or parallel axes to simplify the system of equations. Unlike some other polynomial methods, our polynomial method does not have spurious solutions.

Symbolic manipulation can be performed with Maple \cite{maple} or Mathematica \cite{Mathematica}, although Maple has better performance for polynomial resultants and factorization \cite{howMapleCompares}. A few numerical examples are shown in Appendix~\ref{sec:app_num_examples}.

\subsection{Errors and Subproblems with Tangent Half Angle}
The tangent half angle substitution \(x = \tan(\theta/2)\) results in
\begin{equation} \label{eq:THA_sin_cos}
    \sin(\theta) = \frac{2x}{x^2+1},\quad
    \cos(\theta) = \frac{1-x^2}{x^2+1},
\end{equation}
and the Euler-Rodrigues formula for rotations \cite{euler1776nova, rodrigues1840lois} is
\begin{equation}\label{eq:euler_rodirgues}
    \rot(k,\theta) = k k\tr + \s\theta k^\times  - \c\theta {k^\times}\sq,
\end{equation}
where $k^\times$ denotes the matrix representation of the cross product. For $k=[k_1,k_2,k_3]\tr$,
\begin{equation}
    k^\times := \begin{bmatrix}
    0 & -k_3 & k_2 \\ k_3 & 0 & -k_1 \\ -k_2 & k_1 & 0 
    \end{bmatrix},
\end{equation}
and ${k^\times}\sq = k^\times k^\times =  kk\tr-I$.
Substituting, we get
\begin{equation} \label{eq:rot_THA}
    \rot(k,\theta) =  k k\tr + \frac{2x}{x^2+1} k^\times + \frac{x^2-1}{x^2+1} {k^\times}^2.
\end{equation}

When using \eqref{eq:rot_THA} in the error or subproblem equations, it is important to realize the rotation matrix is orthogonal (\(R\tr R = I\)) but not Hermitian in general (\(R^\dag R \neq I\)). This means the identity \(\norm{Rp} = \norm{p}\) does not hold for complex \(x\) and \(\theta\) since the norm for a complex vector is defined as \(\norm v = \sqrt{v^ \dag v}\). However, we can still use \(p\tr R\tr Rp = p\tr p  = \norm{p}^2\).

We only show how to convert Subproblems~1, 4, and 6 to polynomials as they can be used to solve Subproblems~2, 3, and 5 according to the solutions in Appendix~\ref{sec:app_subproblem_solns}.
For completeness, the tangent half-angle solution to Subproblem~1 is
\begin{equation}
    x = \frac{p_1\tr k^\times p_2}
    {2(p_1\tr k)^2 - p_1\tr p_2 - p_1\tr p_1},
\end{equation}
but the direct rotation solution described below is more computationally efficient.
For Subproblem~4,
 substituting \eqref{eq:rot_THA}, clearing the denominators, and simplifying results in
\begin{equation}
    (2 h\tr k k\tr p - h\tr p - d)x^2
    + (2h\tr k^\times p)x
    + (h\tr p - d) = 0.
\end{equation}
Similarly, for Subproblem~6,
directly substituting \eqref{eq:rot_THA} and clearing the denominators results in a pair of quadrics.
\begin{remark}[Alternative Subproblem Solutions]
    Rather than using this polynomial conversion for robots without three parallel or intersecting axes, these conversions can be used to solve subproblems when all arguments are fixed. For Subproblem 6, we need to eliminate one variable to find a single quartic polynomial in the tangent half angle, and the remaining angle is found by plugging back into the pair of quadrics and finding common zeros (which is equivalent to Subproblem~4).
\end{remark}

\subsection{Direct Rotation Solution for Subproblem 1}
We can simplify the system of polynomials by directly finding \(R(k,\theta)\) for Subproblem~1.
Using the solutions for \(\sin(\theta)\) and \(\cos(\theta)\) in Appendix~\ref{sec:app_subproblem_solns} along with \eqref{eq:euler_rodirgues}, we find
\begin{equation}
    R(k, \theta) = k k\tr + \frac{(k^\times p_1)\tr p_2}{\norm{k^\times p_1}^2} k^\times+
     \frac{({k^\times}^2 p_1)\tr p_2}{\norm{k^\times p_1}^2}{k^\times}^2.
\end{equation}
This can be alternatively derived by writing
\(R(k,\theta) = R_2 R_1 \tr\), where the columns of \(R_i\) are formed by normalizing
\((k,k^\times p_i,-{k^\times}^2 p_i)\).
To avoid the term \(\norm{k^\times p_1}\norm{k^\times p_2}\), as this would add extra radicals to the formulation, we use the substitution \(\norm{k^\times p_1} = \norm{k^\times p_2}\). We can use \(p_1\) or \(p_2\) in the denominator depending on which one is algebraically simpler.

\subsection{Inverse Kinematic for Two Intersecting Joint Axes}
If axes 5 and 6 intersect, the first polynomial in the system comes from the scalar error \eqref{eq:err_2_intersecting}.
Converting \eqref{eq:sp5_two_intersecting} results in two more polynomials and an expression for \(R_{12}\) which is plugged into the error equation.
The result is three equations in three unknowns \(x_1, x_3, x_4\). The resultant can be readily found, which leads to a polynomial in \(x_4\). 
If more than two pairs of axes intersect, the error equation remains the same, but Subproblem~5 may not be needed.

Numerical examples are shown for the FANUC CRX-10iA/L in Appendix~\ref{sec:numerical_CRX} and for the RRC K-1207i with fixed joint 6 in Appendix~\ref{sec:numerical_RRC}.
Simplifications due to intersecting and parallel axes are easy to apply.
For the CRX, the system of polynomials is particularly simple in that there are very few terms and it is highly decoupled, meaning elimination is easier.

\subsection{Inverse Kinematics for General Case}
The vector error \eqref{eq:general6Rerror} is the difference of two unit vectors, meaning there are two degrees of freedom. Projecting the error twice yields two scalar errors which must equal zero and leads to the first two polynomials in the system of equations:
\begin{subequations}
\begin{align} \label{eqn:2d_err_proj_1}
    e_1 &= h_3 \tr R_{34}R_{45} h_6 - h_3 ( R_{01}R_{12})\tr  R_{06}h_6,\\
    e_2 &= h_4 \tr R_{45} h_6 - h_4 ( R_{01}R_{12}R_{23})\tr  R_{06}h_6.
\end{align}
\end{subequations}
Using the Subproblem~5 conversion, \eqref{eq:general6R} turns into a pair of polynomials and an expression for \(R_{34}\), which is plugged into \eqref{eqn:2d_err_proj_1}.
The result is four polynomials in four unknowns \(x_1, x_2, x_3, x_5\).
By eliminating \(x_3\) and \(x_5\), we reduce the systems to two equations in \(x_1, x_2\). We eliminate \(x_2\) to find solutions for \(x_1\), then plug each solution of \(x_1\) into the pair of equations in \(x_1,x_2\) and identify common zeros to find \(x_2\). The remaining joint angles are found in closed form.

There are other options for which joints to eliminate in the errors, Subproblem~5 formulation, and resultant system of polynomials.
Although it is tempting to reformulate \eqref{eqn:2d_err_proj_1} so that the rotation matrix found from Subproblem~1 is not multiplied by another rotation, this is not possible since the eliminated variables cannot be adjacent (here we consider joints 6 and 1 adjacent) to the rotation found with Subproblem~1.

Numerical examples are shown for the Husty \textit{et al.}~\cite{husty2007new} robot in Appendix~\ref{sec:numerical_husty} and for the ABB YuMi with fixed joint 3 in Appendix~\ref{sec:numerical_yumi}. In both cases, subproblem decomposition immediately yields four polynomials from which two joint angles may be found. Since the YuMi has simpler kinematics, the number of terms in the four polynomials is smaller, and the univariate polynomial is of degree 12 rather than the maximum 16.

\subsection{Rational Approximations and Pose Parameterization}
Factoring and finding resultants of polynomials becomes more efficient when coefficients are rational numbers as opposed to radical expressions or symbolic sines and cosines of angles. 
We can find arbitrarily accurate rational approximations for \(p_{i,i+1}\), \(h_i\), and \(R_{06}\) to speed up computation, opening up a tradeoff between approximation accuracy and computational speed.
Rational approximations for each \(p_i\) is straightforward, although computational efficiency is improved by maintaining cases of intersecting, orthogonal, or parallel axes or link offsets.

The rotation \(R_{06}\) only appears as multiplying another vector; for example, \(R_{06} h_6\). We can parameterize \(R_{06}\) as the product of three consecutive orthogonal rotations \(R_{\gamma}R_{\beta}R_{\alpha}\). Picking \(R_{\alpha}=\rot(h_6, \alpha)\) means \(\alpha\) does not appear in the equation. Similarly, each \(h_i\) can be written as a series of rotations applied to some basis vector.
The rational approximation for \(R_{06}\) and \(h_i\) can then be reduced to finding rational approximations for \(\tilde s \approx \sin (\theta)\) and \(\tilde c \approx \cos (\theta)\), while maintaining the identity \({\tilde s}^2 + {\tilde c}^2 = 1\). This can be done by finding the rational approximation for \(\tan(\theta/2)\) and plugging into \eqref{eq:THA_sin_cos} or \eqref{eq:rot_THA}. This is equivalent to finding Pythagorean triples which are the side lengths of a right triangle with one angle close to \(\theta\).
Parameterizing the end effector pose in this way means the polynomial system can be found a priori in terms of \(p_{0T}, \beta, \gamma\).

\subsection{Comparison to Search Method}
The polynomial method has several advantages over the search method.
Finding the univariate polynomial is straightforward
    since most of the algebra is performed automatically.
Polynomial root finding is a solved problem, so
    it is guaranteed to find all solutions including complex solutions,
    and repeated roots can be detected as robot singularities.

Despite the utility of the polynomial method, the search method remains a more practical approach.
Although the system of polynomials only needs to be derived once,
    each new end effector pose requires symbolic manipulation to find a new resultant univariate polynomial.
This means the search method is much more computationally efficient,
    even with the somewhat rudimentary search implementation we present here.
The tangent half angle substitution also leads to issues with joint angles around $\pm\pi$,
    including reduced accuracy or polynomial deflation \cite{fundamentalsBook},
    although there are strategies to overcome this singularity issue,
    such as by running the algorithm with two different zero configurations.

The polynomial and search methods are each useful for different reasons.
Whereas the search method is fast and suitable for real-time control,
    the polynomial method provides stronger guarantees on the IK solutions returned
    and is more suited for analysis of a few end effector poses at a time.

\section{Conclusion}\label{sec:conclusion}
This paper revisits the subproblem decomposition method for solving robot inverse kinematics.
By converting a nonlinear problem to a linear one and then imposing constraints on the linear solutions, we found closed-form solutions to the original three Paden--Kahan subproblems as well as three additional subproblems. For Subproblems~1--4, we found the least-squares solutions in closed form when the exact problem has no solution, and we similarly found continuous approximate solutions for Subproblems~5 and 6.
The inverse kinematics for any robot with three consecutive intersecting or parallel axes may be decomposed into a series of subproblems and solved directly in closed form.  For other robots, the decomposition method reduces the solution to a 1D or 2D search. This method finds {\em all} inverse kinematics solution instead of just one solution close to an initial guess as in Jacobian-based methods. We have also shown how to extend the subproblem decomposition method to find a high-order polynomial in the tangent half angle of one of the joints.

We have demonstrated excellent computational performance for IK using the subproblem decomposition method, and there are still ways to speed up computation time. The implementation code for most IK solutions can be better optimized as most of computation time is spent not on solving subproblems but on other calculations such as computing rotation matrices. Perhaps performance can be improved by computing rotations using quaternions or by directly using the sine and cosine from each subproblem solution. For search-based IK solutions, there are a number of possible algorithmic improvements. Sampling could be parallelized, or a much faster search method could be used, such as one that immediately starts optimization after just a few random samples, doing a random restart if that optimization fails. It may also improve performance to increase sampling in regions next to the boundary of the domain or in regions that may be local minima or maxima pointing towards zero. For 2D search, using information from each element of the vector error, rather than the norm, may also improve performance.
The connection to the tangent half angle polynomial method demonstrates that we can easily find a system of three or four polynomials, but computational performance may be improved with more efficient elimination methods. Although finding the rotation matrix directly for Subproblem~1 helps, it may be possible to find a more compact polynomial for the error equation. 

There are applications for these subproblems beyond just inverse kinematics for 6R arms in other geometric problems such as inverse kinematics of robots with prismatic joints, forward and inverse kinematics for parallel manipulators, or certain problems found in computer graphics and animation.

% \appendix
\appendices
\ifabridgedsolutions
\input{abridged_sp_solns}
\else
\section{Subproblem Solutions} \label{sec:app_subproblem_solns}
We solve Subproblems~2 and 3 by reducing them to Subproblem~4, and we solve Subproblem~5 by manipulating the Subproblem~4 solution. We solve the remaining Subproblems~1, 4, and 6 by writing them as a linear equations in \(x = \begin{bmatrix}\s\theta&  \c\theta\end{bmatrix}\tr\), where $\s\theta=\sin\,\theta$ and $\c\theta=\cos\,\theta$, with nonlinear constraint \(\sin^2(\theta) + \cos^2(\theta) = \norm{x}^2 = 1\). We first solve the unconstrained linear problem with the tools of linear algebra, and then find the subset of solutions following the constraint.
A similar approach was used in \cite{norrdine2012algebraic} to find intersections among spheres.
To convert a subproblem to a linear problem, we may rewrite \eqref{eq:euler_rodirgues} as
\begin{equation} \label{eq:rotation_as_linear}
    \rot(k,\theta)p = p_k + A_{k,p} x,
\end{equation}
where 
\begin{equation}
    p_k = k k\tr p
    \quad \text{and} \quad
    A_{k,p} = \begin{bmatrix}
     k^\times p & - {k^\times}\sq p
    \end{bmatrix}.
\end{equation}
Geometrically, this conversion replaces any circle with the plane in which the circle lies. The vector \(p_k\) is along the circle axis pointing to the circle center, and the columns of \(A_{k,p}\) form an orthogonal basis for the plane and have norm equal to the circle radius.

In Subproblems~1--4, if an exact solution does not exist, then we instead find the solution which minimizes Euclidean distance. Conveniently, this least-squares solution requires almost no overhead to compute. The minimizing solutions between two smooth objects occurs on a line normal to both objects. As pointed out in \cite{zsombor2004extreme}, for a circle the common normal passes through the circle axis. Similarly, in Subproblems~5 and 6, we find continuous approximate solutions by using the real part of any complex pairs of roots of the quartic polynomial.

Given a solution for \(x\), the corresponding angle is found using \(\theta = \mbox{ATAN2}(x_1,x_2)\). ATAN2 is the only inverse trigonometric function used as it is extremely robust compared to \(\sin^{-1}()\) and \(\cos^{-1}()\) \cite{diankov_thesis},
as well as compared to solving for \(\tan(\theta/2)\) \cite{manocha1994efficient, fundamentalsBook}. We can achieve some computational speed-up by recognizing \(\mbox{ATAN2}(\s\theta,\c\theta) = \mbox{ATAN2}(\alpha \s\theta,\alpha \c\theta)\) where \(\alpha > 0\). This means we only have to solve for \(\alpha x\).

\begin{subsectionbox}{Subproblem 1: Circle and Point}
Given vectors $p_1$, $p_2$ and a unit vector \(k\), find $\theta$
to minimize $\norm{\rot(k,\theta)p_1-p_2}$.
\end{subsectionbox}
Assume $k$ and $p_1$ are not collinear, and assume  $k$ and $p_2$ are not collinear. Otherwise, $\theta$ is arbitrary.  Using \eqref{eq:rotation_as_linear}, we have
\begin{align}
    \norm{\rot(k,\theta)p_1-p_2}  = \norm{A_{k,p_1} x - b},
    \label{eq:subprob1app0}
\end{align}
where \(b = p_2 - {p_1}_k = p_2 - k k\tr p_1\).
By assumption, \(A_{k,p_1}\) has full column rank. An exact solution to \(A_{k,p_1}x = b\) such that \(\norm{x} = 1\) exists when \(p_2\) lies on the circle, meaning \(\norm{p_1} = \norm{p_2}\) and \(k\tr p_1 = k\tr p_2\). In this case, \(x\) is the unconstrained least-squares solution
\begin{equation}
x_{LS}=  \min_x\norm{A_{k,p_1}x-b} = A_{k,p_1}^+ b = A_{k,p_1}^+ p_2,
\label{eq:subprob1solution}
\end{equation}
where \(A^+ =  (A\tr A)^{-1} A\tr\) is the pseudo-inverse (left inverse) of \(A\). One can show \(A_{k,p_1}^+ = {A_{k,p_1}\tr}/{\norm{k^\times p_1}\sq}\).

In general, an exact solution is not possible, and we need to solve the constrained least-squares problem $\min_x\norm{A_{k,p_1} x - b}$ subject to $\norm{x}=1$. Since the common normal between the circle and \(p_2\) must pass through the circle's axis, the minimizing solution occurs on the plane of symmetry containing the circle's axis and \(p_2\).
As shown in Fig.~\ref{fig:subprob1pic}, \(x_{LS}\) is the coordinates in the coordinate system \((-{k^\times}^2 p_1, k^\times p_1)\) of $p_2$ projected onto the plane containing the circle. This vector is in the plane of symmetry, and normalizing it gives the constrained least-squares solution \(x = x_{LS} / \norm{x_{LS}}\).
Skipping all positive scalar division (which keeps the value of ATAN2 unaffected) yields the general solution
\begin{equation}
    \theta = \mbox{ATAN2}( (k^\times p_1)\tr p_2,- ({k^\times}\sq p_1)\tr p_2).
    \label{eq:subprob1soltheta}
\end{equation}
Note that all other \(p_2\) vectors which lie in the same half plane containing \(k\) result in the same least-squares solution.

\begin{figure}[tb]
    \centering
    \includegraphics[scale=0.5, clip]{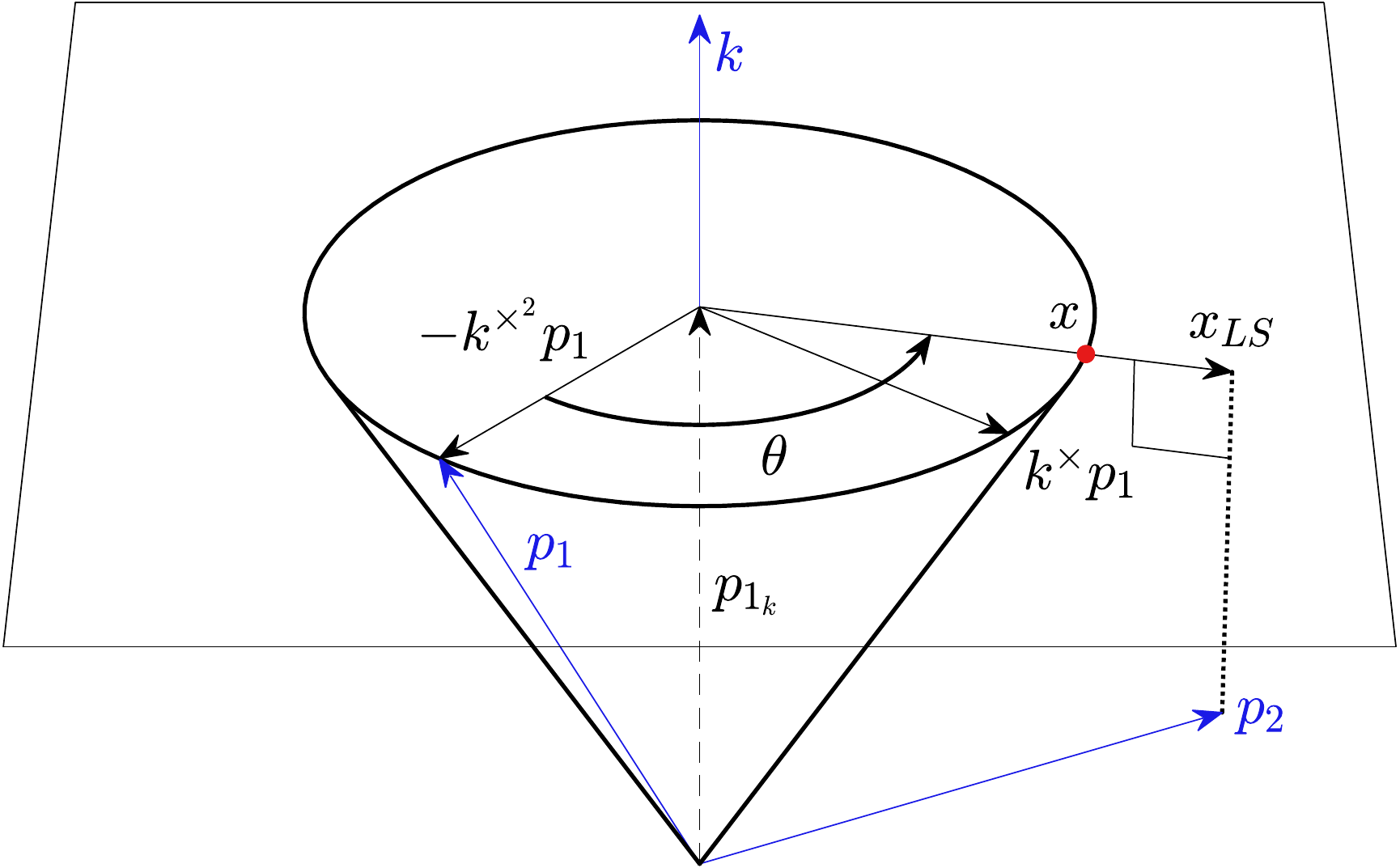}
    \caption{Subproblem 1 finds the angle of the closest point on the circle \(\rot(k,\theta)p_1\) to the point \(p_2\). This is the angle of \(x_{LS}\) in the coordinate frame in the circle plane formed by \((-{k^\times}^2 p_1, k^\times p_1)\), where \(x_{LS}\) is the projection of \(p_2\) to the plane containing the circle.}
    \label{fig:subprob1pic}
\end{figure}

% \pagebreak[3]
\begin{subsectionbox}{Subproblem 2: Two Circles}
Given vectors $p_1$, $p_2$ and unit vectors $k_1$, $k_2$, find $(\theta_1,\theta_2)$
to minimize $\norm{\rot(k_1,\theta_1) p_1-\rot(k_2,\theta_2)p_2}$.
\end{subsectionbox}

Assume the following pairs of vectors are not collinear: 
$(k_1,p_1)$, $(k_2,p_2)$, $(k_1,k_2)$.  Otherwise, the problem reduces to Subproblem 1 and \(\theta_1\), \(\theta_2\), or \(\theta_1 \pm \theta_2\) are arbitrary, respectively.

An exact solution \(\rot(k_1,\theta_1) p_1 = \rot(k_2,\theta_2)p_2\) is possible only if \(\norm{p_1} = \norm{p_2}\), as this means the two circles lie on the same sphere. We see the minimizing \((\theta_1, \theta_2)\) is independent of \(\norm{p_1}\) and \(\norm{p_2}\) if we rewrite the subproblem as
\begin{multline}
    \norm{\rot(k_1,\theta_1) p_1-\rot(k_2,\theta_2)p_2}^2 \\
    = \norm{p_1}^2 + \norm{p_2}^2 -2\norm{p_1} \norm{p_2} \cos(\psi),
\end{multline}
where \(\psi\) is the angle between the rotated vectors. This means we can find the least-squares solution(s) by normalizing \(p_1\) and \(p_2\) before proceeding.
Once we assume \(p_1\) and \(p_2\) are the same length, an exact solution exists if and only if
\begin{equation}
    \abs{\theta_{k_1,p_1} - \theta_{k_2,p_2}}
        \leq \theta_{k_1,k_2}
        \leq \theta_{k_1,p_1} + \theta_{k_2,p_2},
\end{equation}
where \(\theta_{v_1,v_2}\) is the angle between vectors \(v_1\) and \(v_2\).
This inequality can be derived by considering the intersections of the circles with the plane of symmetry spanned by \(k_1\) and \(k_2\). A solution exists if and only if the intersections are interleaved, that is, the intersections alternate between circles 1 and 2 as we rotate along the plane of symmetry.
If \(p_1\) is arbitrary, then \(\theta_{k_1,p_1}\) can be any angle from \(0\) to \(\pi\), and a solution always exists if and only if \(k_1\tr k_2 = 0\) and \(k_2\tr p_2 = 0\).

Projecting the Subproblem~2 equation onto \(k_2\) or \(k_1\) gives equations only in terms of \(\theta_1\) or \(\theta_2\) which can be solved using Subproblem~4:

\begin{subequations}
\begin{align}
    k_2 \tr \rot(k_1,\theta_1) p_1 &= k_2 \tr p_2,\\
    k_1 \tr \rot(k_2,\theta_2) p_2 &= k_1 \tr p_1.
\end{align}
\end{subequations}
If there are two solutions for \(\theta_1\) and \(\theta_2\), ensure the solutions match by switching the order of solutions for \(\theta_2\).
Sometimes \(\theta_1\) and \(\theta_2\) may have different numbers of solutions due to numerical issues; in this case, duplicate the solution for the angle with only one solution.
Subproblem~4 finds the intersections between each circle and the plane in which the other circle lies, as shown in Fig.~\ref{fig:subprob2}(a). Since the two circles lie on the same sphere. and there are no extraneous solutions: each intersection between a circle and the other plane must also be an intersection with the other circle.

If the two circles do not intersect, as in Fig.~\ref{fig:subprob2}(b), the least-squares solution is also given by Subproblem~4, and this case can be detected by checking if Subproblem~4 returns a least-squares result. The minimizing solutions occur when line passing through \(\rot(k_1,\theta_1) p_1\) and \(\rot(k_2,\theta_2) p_2\) passes through both circle axes, meaning the solutions must occur on the plane of symmetry containing \(k_1\) and \(k_2\). (We have removed the case of the line being out of plane but passing through the origin by assuming \(\norm{p_1} = \norm{p_2}\).) The least-squares solution for Subproblem~4 lies on this same plane.

\begin{remark}[Alternate Decomposition]
Alternatively, solve Subproblem 4 for one angle, then use Subproblem~1 to find the remaining angle solutions.
\end{remark}

\begin{remark}[Linear Solution]
This subproblem may be solved by rewriting the problem as a linear equation \(A x = b\), where \begin{equation}\label{eq:subprob2_pAx}
A = \begin{bmatrix}
    A_{k_1,p_1} &A_{k_2,p_2}
\end{bmatrix} \text{ and }
b = {p_2}_k - {p_1}_k.
\end{equation}
The complete solution \( x = x_{min} + x_N\) is the intersection between the two planes, where \(x_{min}\) is the minimum-norm solution and \(x_N\) is an arbitrary vector in the one-dimensional null space of \(A\). Simplifying to find efficient expressions for \(x\) such that \(\norm{x_i} = 1\) results in the solution above.
\end{remark}

\begin{figure}[t]
    \centering
    \subfloat[]{\includegraphics[scale=0.5, clip]{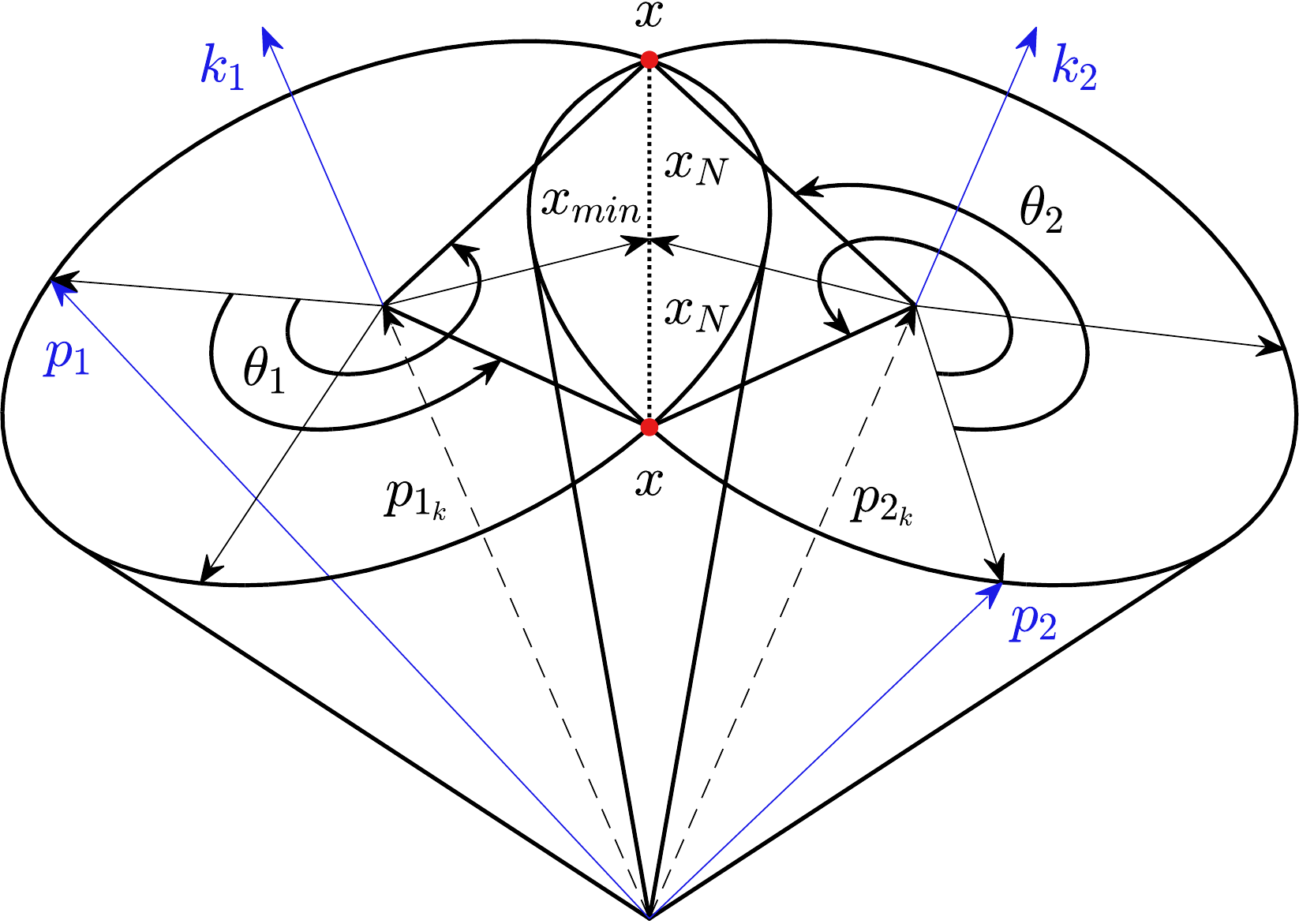}}\\
    \subfloat[]{\includegraphics[scale=0.5, clip]{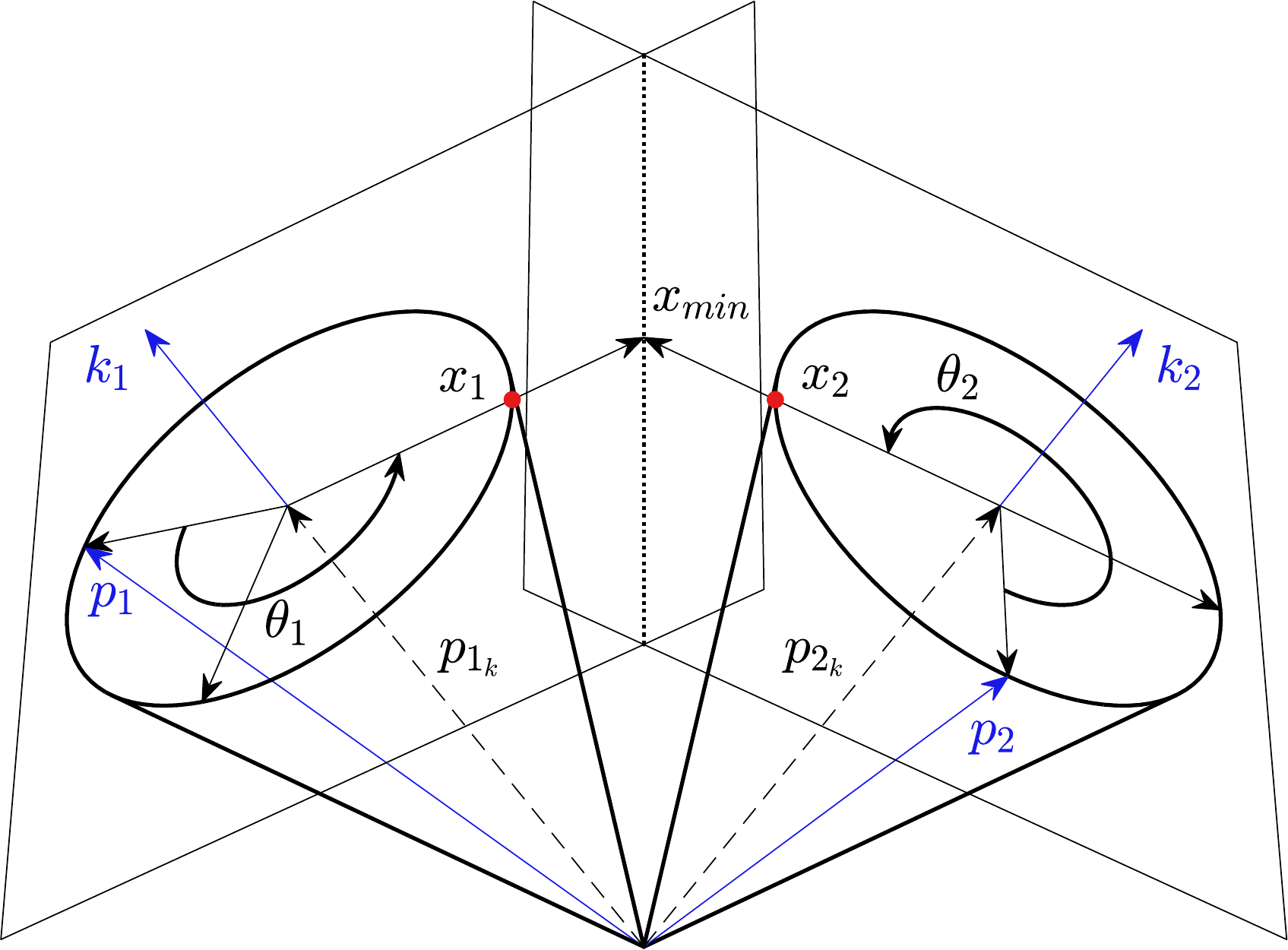}}
    \caption{Subproblem 2 finds the angles of the closest points on the two circles \(\rot(k_1,\theta_1)p_1\) and \(\rot(k_2,\theta_2)p_2\).  (a) Two intersecting circles with two intersection points. The solutions are found along the intersection of the planes containing the circles. (b) Two non-intersecting circles. The minimizing solution is the angle of the minimum-norm point on the intersection of the two planes containing the circles.}
    \label{fig:subprob2}
\end{figure}

\begin{remark}[Skew Axes]
It is straightforward to extend this subproblem to the case where the circle axes are skew, solving
\begin{equation}
    p_0+\rot(k_1,\theta_1) p_1 = \rot(k_2,\theta_2)p_2.
\end{equation}
The solution is nearly identical: Project onto \(k_1\) or \(k_2\) and solve Subproblem~4 and Subproblem~1 without normalizing \(p_1\) and \(p_2\). Since the axes are skew, there is only up to one solution, and we need to check for extraneous solutions by checking if both subproblems returned an exact solution. This problem can be solved in a similar way using Subproblem~3 rather than Subproblem~4 by taking the norm of both sides of the equation. Solving the least-squares version of this problem is possible but complicated as there are up to eight common normals between two circles \cite{odehnal2005common}.

When the circle axes are skew, the two circles lie on different spheres, and the problem is generally unsolvable: with randomly chosen parameters, there is almost surely no solution for \((\theta_1, \theta_2)\). It is therefore inadvisable to use this as part of the solution to a larger geometry problem.
\end{remark}

\begin{remark}[Intersecting Axes]
If \(p_0\) can be written as a linear combination of \(k_1\) and \(k_2\), then the circle axes still intersect, and the problem can be solved with Subproblem~2 by moving the respective components of \(p_0\) to \(p_1\) and \(p_2\).
\end{remark}

\begin{remark}[Parallel Axes]
If the circle axes are parallel ($k_1=k_2=k$), we require $k\tr(p_0+p_1-p_2)=0$ to guarantee exact solutions. Up to two solutions of $\theta_1 $ may be solved using Subproblem~3, and the corresponding solutions of $\theta_2$ may be found using Subproblem~1.
We can also find up to two least-squares solutions if $k\tr(p_0+p_1-p_2)\neq0$ by replacing \(p_i\) with \(-{k^\times}^2p_i\) for \(i = 0,1,2\).
\end{remark}

% \pagebreak[3]
\begin{figure}[t]
    \centering
    \subfloat[]{\includegraphics[scale=0.5, clip]{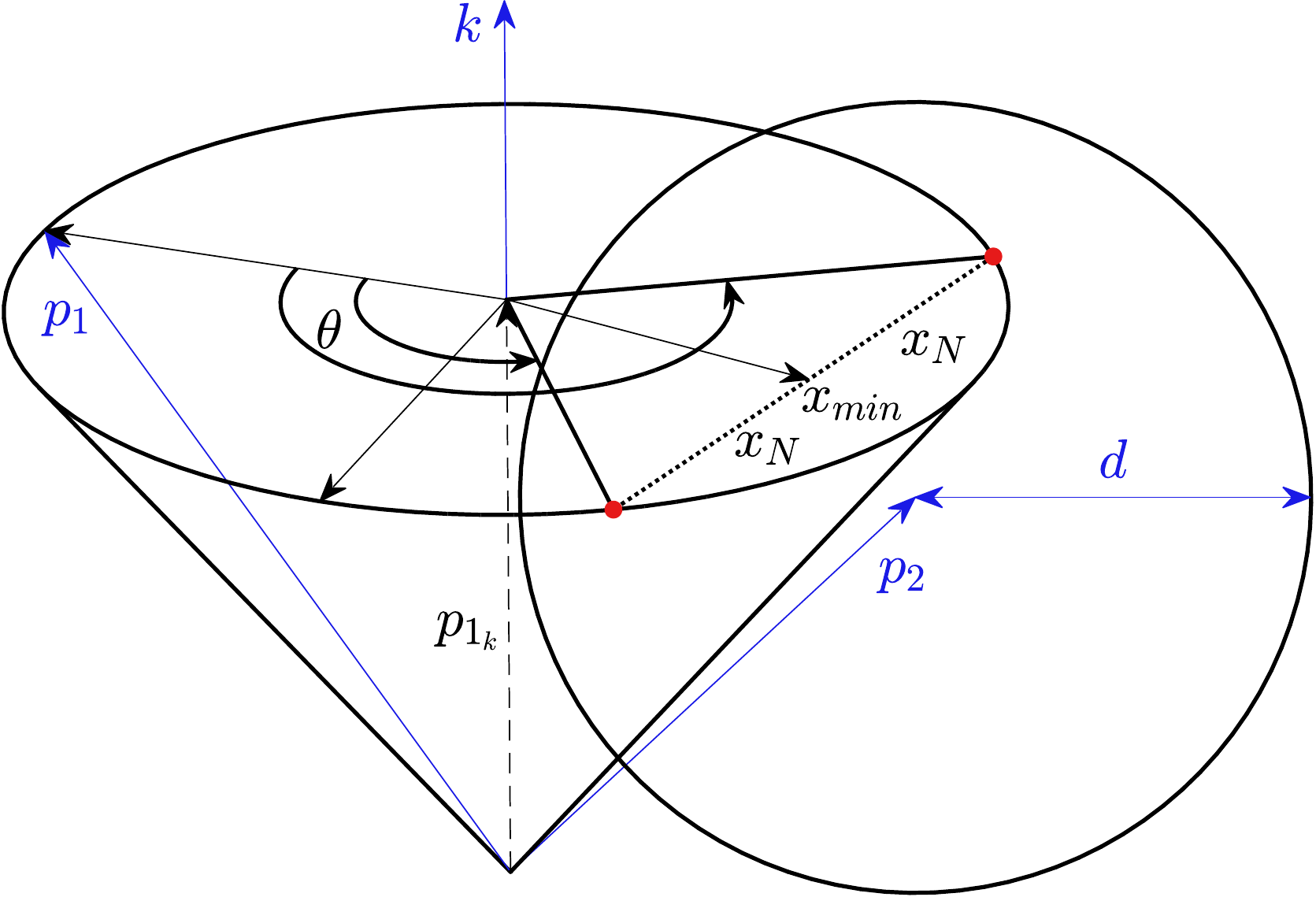}}\\
    \subfloat[]{\includegraphics[scale=0.5, clip]{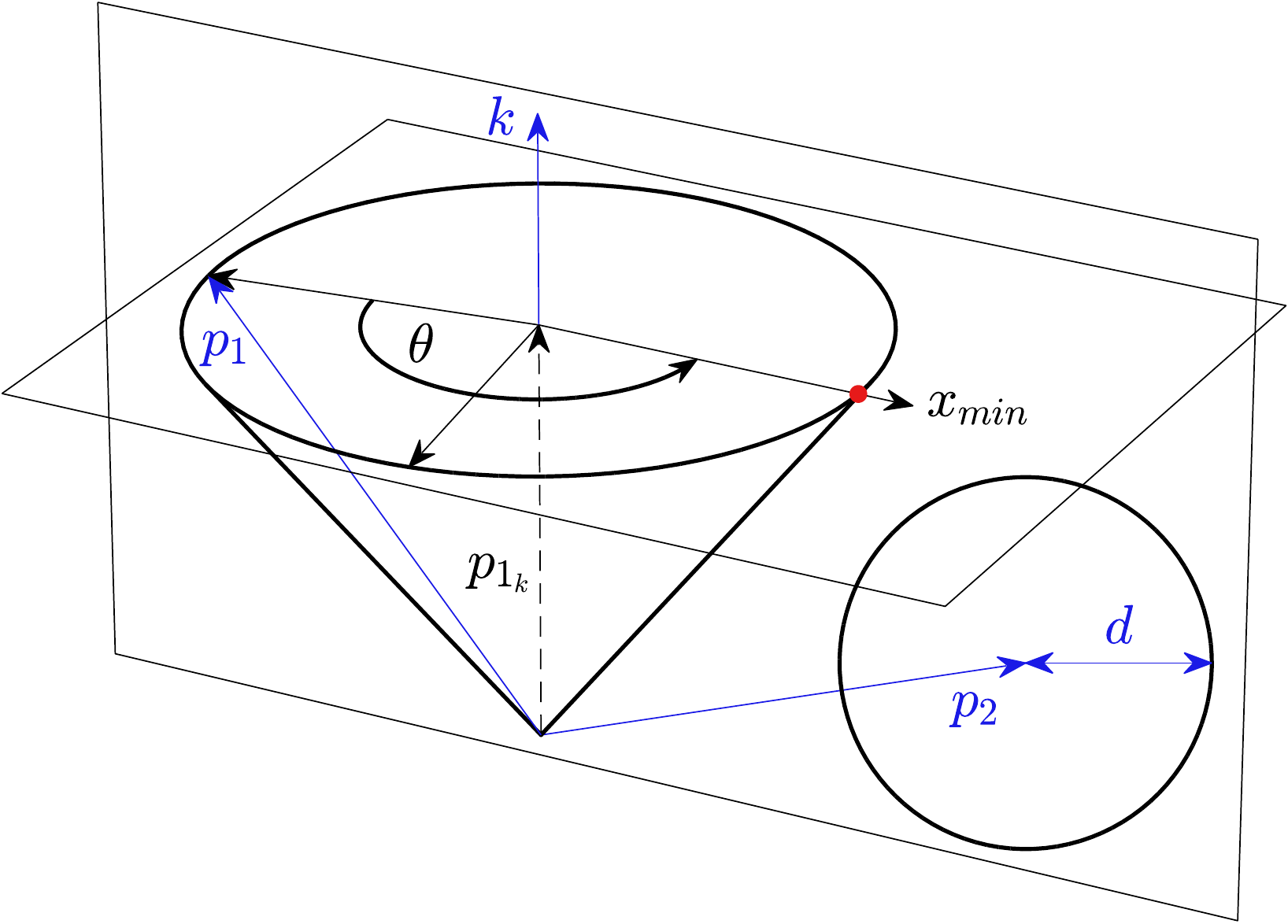}}
    \caption{Subproblem 3 finds the angles of the closest points on the circle \(\rot(k,\theta)p_1\) to the sphere of radius \(d\) centered at \(p_2\).  (a) Intersecting circle and sphere. (b) Minimum distance between a non-intersecting circle and sphere.}
    \label{fig:subprob3pic}
\end{figure}
\begin{subsectionbox}{Subproblem 3: Circle and Sphere}
Given vectors $p_1$, $p_2$, a unit vector $k$, and a nonnegative scalar $d$, find $\theta$
to minimize $\abs{\norm{\rot(k,\theta)p_1-p_2}-d}$.
\end{subsectionbox}

Assume $k$ and $p_1$ are not collinear, and assume \(k\) and \(p_2\) are not collinear; otherwise, the problem is independent of $\theta$. 

Write the problem equivalently as minimizing
\begin{multline}\label{eq:sp3_equiv}
    \abs{ \frac{1}{2} \norm{\rot(k,\theta)p_1-p_2}\sq - \frac{d\sq}{2} }
    = \\
    \abs{
        p_2 \tr \rot(k,\theta) p_1 - \frac{1}{2}(
        \norm{p_1}^2
        +\norm{p_2}^2
        -d^2
        )
    },
\end{multline}
which can be solved using Subproblem~4. If the circle and sphere intersect, as in Fig.~\ref{fig:subprob3pic}(a), then this conversion, which is basically the law of cosines, replaces the sphere with a plane which intersects the circle at the same points. If the circle and sphere do not intersect, as in Fig.~\ref{fig:subprob3pic}(b), the minimizing angle on the circle to the plane is also the minimizing angle to the sphere, which occurs on the line passing through the circle axis and the center of the sphere. Interestingly, if \(d=0\) then this subproblem solves Subproblem~1.

% \pagebreak[3]
\begin{subsectionbox}{Subproblem 4: Circle and Plane}
Given a vector $p$, unit vectors $k$, $h$, and a scalar $d$, find $\theta$ to minimize $\abs{h\tr\rot(k,\theta)p-d}$.
\end{subsectionbox}
Assume $p$ and $k$ are not collinear, and assume $h$ and $k$ are not collinear; otherwise, the problem does not depend on $\theta$. If we apply \eqref{eq:rotation_as_linear}, then
\begin{equation}
    h\tr\rot(k,\theta)p-d
    = A x - b, 
\end{equation}
\begin{figure}[tb]
    \centering
    \subfloat[]{\includegraphics[scale=0.5, clip]{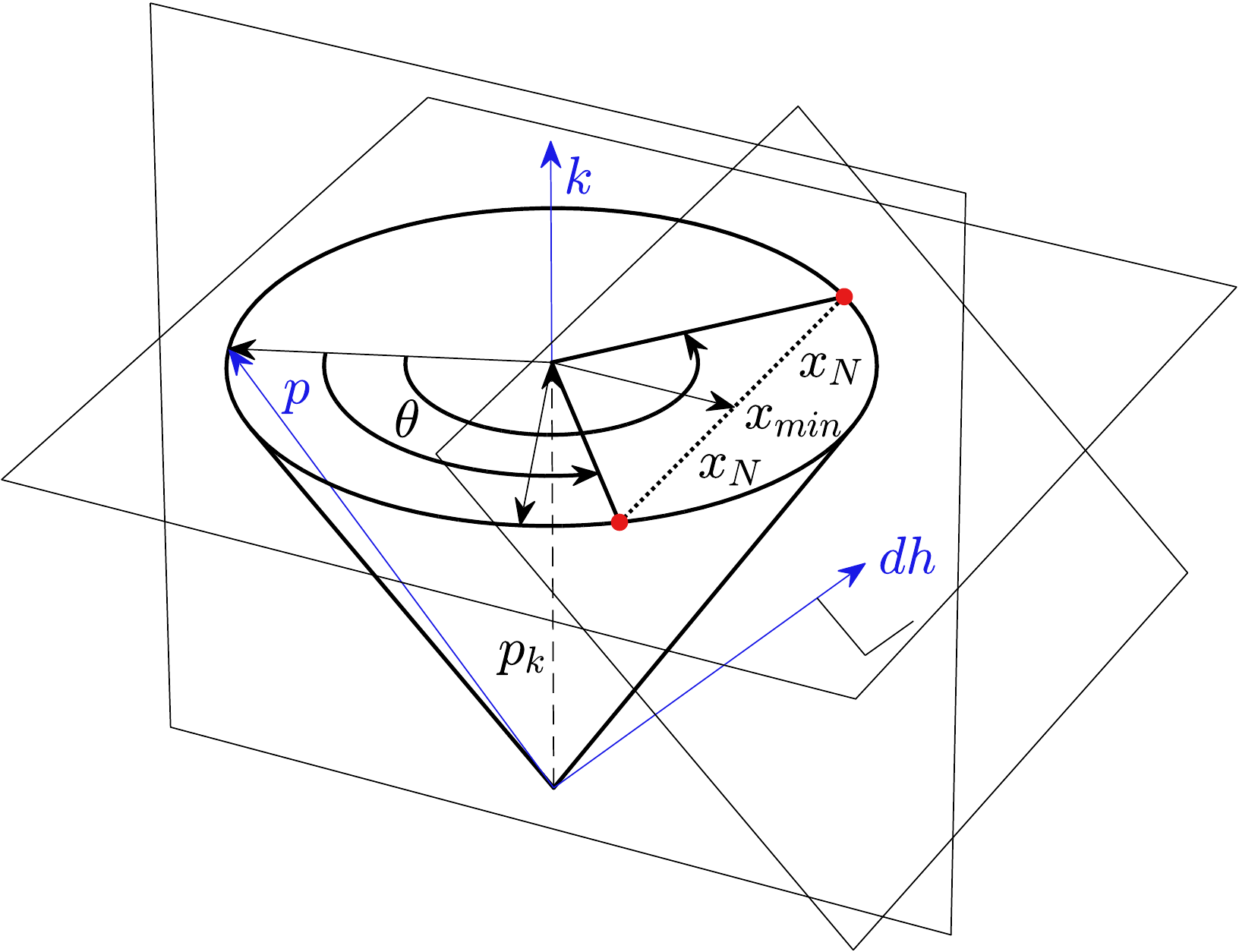}}\\
    \subfloat[]{\includegraphics[scale=0.5, clip]{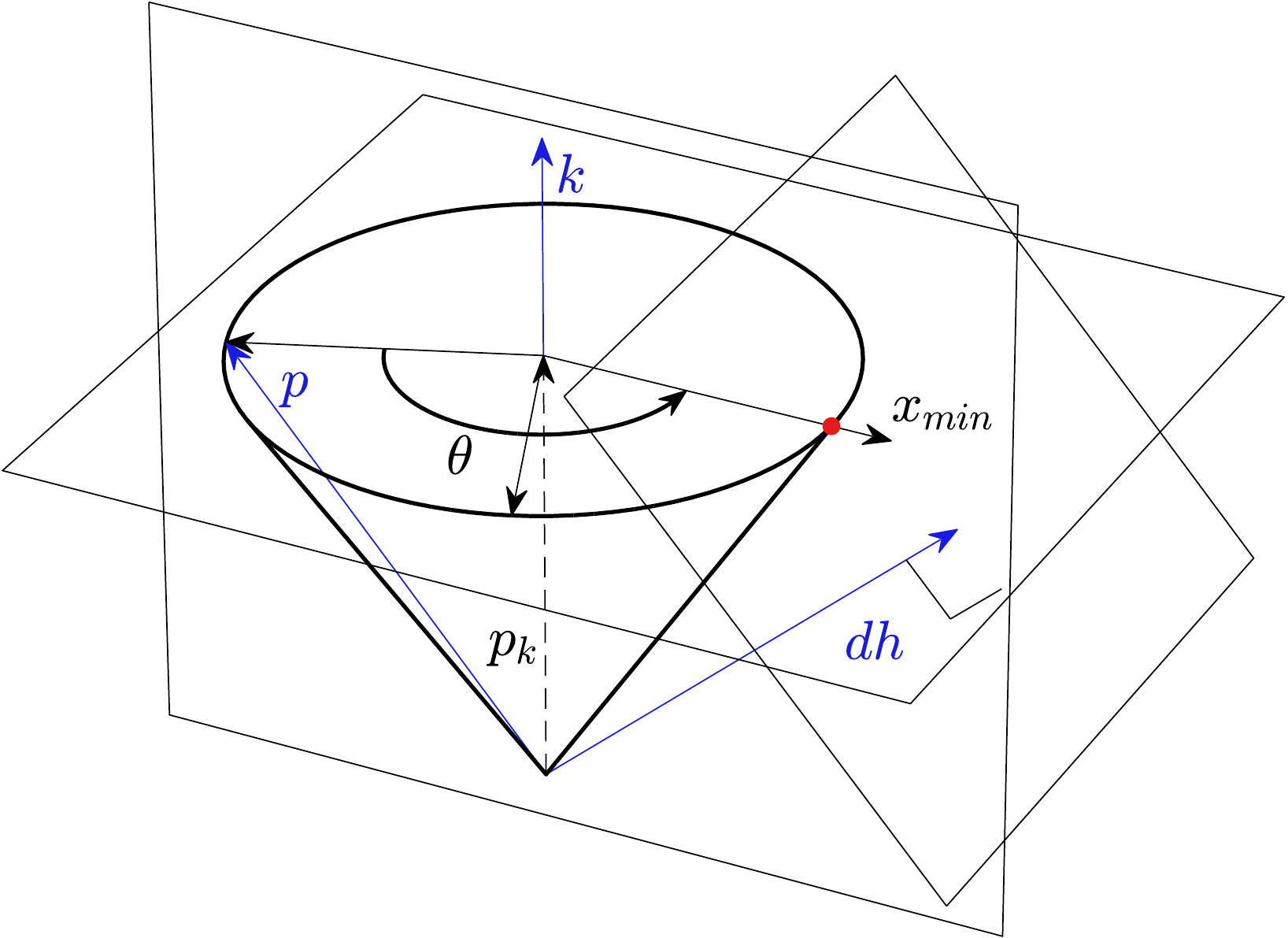}}
    \caption{Subproblem 4 finds the angles of the closest points on the circle \(\rot(k,\theta)p\) to the plane with normal vector \(h\) and distance \(d\) from the origin.  (a) A circle intersecting with a shifted plane. (b) Minimum distance between non-intersecting circle and plane.}
\label{fig:subprob4pic}
\end{figure}
where \(A =h\tr A_{k,p}\) and \(b = d-h\tr k k\tr p\).
There are many solutions to \(Ax=b\), which can be parameterized as \(x = A^+b + x_N\), where \(A^+ = A\tr (A A\tr)^{-1}\) (for $A$ with full row rank) is the pseudo-inverse (right inverse) of \(A\),  \(x_{\min} = A^+b\) is the minimum-norm solution, and \(x_N\) is an arbitrary vector in the null space of \(A\), which is orthogonal to \(x_{min}\).
By assumption, the $2\times1$ row vector $A$ is of rank one and has a one-dimensional null space given by
\begin{equation} \label{eq:sp3_null_space}
    x_N= \xi\,\, x_N',\quad x_N'= \underbrace{\ma{cc} 0 & 1 \\ -1 & 0\ema}_J A\tr,
\end{equation}
where \(x'_{N}\) forms a basis for the null space and \(\xi\) parameterizes the solution.
We can use \(\norm{x}^2 = 1\) to find \(\xi\) as the solution to a quadratic equation.
The null space is the intersection between the plane containing the circle and a plane perpendicular to $h$.  The minimum-norm solution $x_{min}$ is given by 
\begin{equation} \label{eq:sp3_x_min}
    x_{min} = \frac{A\tr }{\norm{A\tr}\sq}\, b,
\end{equation}
which is the point of intersection among three planes: The plane containing the circle, the plane perpendicular to \(dh\), and the plane of symmetry spanned by $k$ and $h$.

If $\norm{x_{min}} < 1$, as in Fig.~\ref{fig:subprob4pic}(a), two intersections occur, \(x_{min}\) corresponds to an interior point of the circle, and the two choices for \(\xi\) are
\begin{equation} \label{eq:sp3_xi}
    \xi = \pm\frac{\sqrt{1-\norm{x_{min}}\sq}}{\norm{x_N'}}
   = \pm\frac{\sqrt{\norm{A\tr}\sq-b\sq}}{\norm{A\tr}\sq}.
\end{equation}
If \(\norm{x_{min}} = 1\), one intersection occurs, and \(x = x_{min}\). 
 If $\norm{x_{min}}>1$, as in Fig.~\ref{fig:subprob4pic}(b), the circle does not intersect with the plane. The minimizing solution occurs on the common normal which is perpendicular to the plane and passes through the circle axis. This means the solution occurs on the plane of symmetry, and \(x\) is therefore the normalized version of \(x_{min}\). We can skip the normalization since we are plugging \(x\) into ATAN2.

 If \(h\) is not a unit vector, the solution method stays the same. The only difference is in the geometric interpretation.

% \pagebreak[3]
\begin{subsectionbox}{Subproblem 5: Three Circles}
Given vectors, $p_0$, $p_1$, $p_2$, $p_3$ and unit vectors $k_1$, $k_2$, $k_3$, find $(\theta_1,\theta_2,\theta_3)$
to solve
\begin{equation}  \label{eq:subproblem_5}
p_0+\rot(k_1,\theta_1) p_1 = \rot(k_2,\theta_2)(p_2+\rot(k_3,\theta_3)p_3).
\end{equation}
\end{subsectionbox}

We may visualize this subproblem as the intersection of three circles, as depicted in Fig.~\ref{fig:subprob6pic}: 
\begin{itemize}
    \item Circle 1 is given by $p_0+\rot(k_1,\theta_1)p_1$.
    \item Circle 2 has axis $k_2$. Its radius \(r\) and height \(z\) are to be determined, where \(z\) may be negative if the circle center in the \(-k_2\) direction.
    \item Circle 3 is given by $p_2+\rot(k_3,\theta_3)p_3$.
\end{itemize}

Assume \((k_1, p_1)\) and \((k_3, p_3)\) are not collinear, as otherwise \(\theta_1\) or \(\theta_3\) is arbitrary. (Other cases for a continuum of solutions exist if \(p_0\tr k_1^\cross k_2=0\) or \(p_2\tr k_3^\cross k_2 =0 \).)
{The projection of the solutions on circle 1 and circle 3 to $k_2$ is \(z\):
\begin{equation}
    z= k_2\tr(p_0+\rot(k_1,\theta_1)p_1)
    = k_2\tr(p_2+\rot(k_3,\theta_3)p_3).
\end{equation}
Write 
\begin{subequations}
\begin{align}
   p_0+\rot(k_1,\theta_1)p_1 &= A_{k_1,p_1} x_1 + p_{1_S},\\
   p_2+\rot(k_3,\theta_3)p_3 &= A_{k_3,p_3} x_3 + p_{3_S},
\end{align} \label{eq:sp5_circles_linear}
\end{subequations}
where the shifted centers of the circles are
\begin{equation}
p_{1_S}=p_0+ k_1k_1\tr p_1, \quad p_{3_S}=p_2+ k_3k_3\tr p_3.
\end{equation}
Use Subproblem 4 to find $x_1$ and $x_3$ in terms of $z$:
\begin{equation}
    x_i =
    \frac{v_i (z-\delta_i) \pm J v_i \sqrt{\norm{v_i}\sq-(z-\delta_i)\sq}}{\norm{v_i}\sq},
\label{eq:x1_x3}
\end{equation}
where \(J\) is defined in \eqref{eq:sp3_null_space}, \(v_i = A_{k_i,p_i}\tr k_2\), and \(\delta_i =  k_2\tr p_{i_S}\).
By substituting \eqref{eq:x1_x3} into \eqref{eq:sp5_circles_linear} and taking the squared norm we obtain equations for two ellipses in \((r^2 + z^2, z)\), which are distorted versions of the cross sections of the surfaces of revolution of circles 1 and 3 around \(k_2\):
\begin{align}\label{eq:subprob5radius}
\begin{split}
    r^2 + z^2 
    =&{} \norm{k_i^\cross p_i }^2 + \norm{p_{i_s}}^2 + 2 \alpha_i \delta_i  - 2 \alpha_i z\\
    & \pm 2 \beta_i
    \sqrt{\norm{k_i^\cross p_i }^2\norm{k_i ^\cross k_2}^2 - (z - \delta_i)^2},
\end{split}
\end{align}
where
\begin{equation}
    \alpha_i = \frac{p_{i-1}\tr {k_i^\cross}\sq k_2}{\norm{k_i^\cross k_2 }^2}
\text{ and }
    \beta_i = \frac{p_{i-1}\tr k_i^\cross k_2}{\norm{k_i^\cross k_2 }^2}.
\end{equation}
To arrive at \eqref{eq:subprob5radius}, we have used
\begin{align}
    A_{k_i,p_i} J A_{k_i,p_i}\tr &= \norm{k_i^\times p_i}^2 k_i ^\times,\\
    \norm{A_{k_i,p_i}\tr k_2}^2 &=  \norm{k_i^\times p_i}^2 \norm{k_i^\times k_2}^2,\\% \intertext{and}
     A_{k_i,p_i} A_{k_i,p_i} \tr &= -\norm{k_i ^\cross p_i}^2 {k_i^\cross}\sq.
\end{align}
\begin{figure}[tb]
\centering
\includegraphics[scale=0.5, clip]{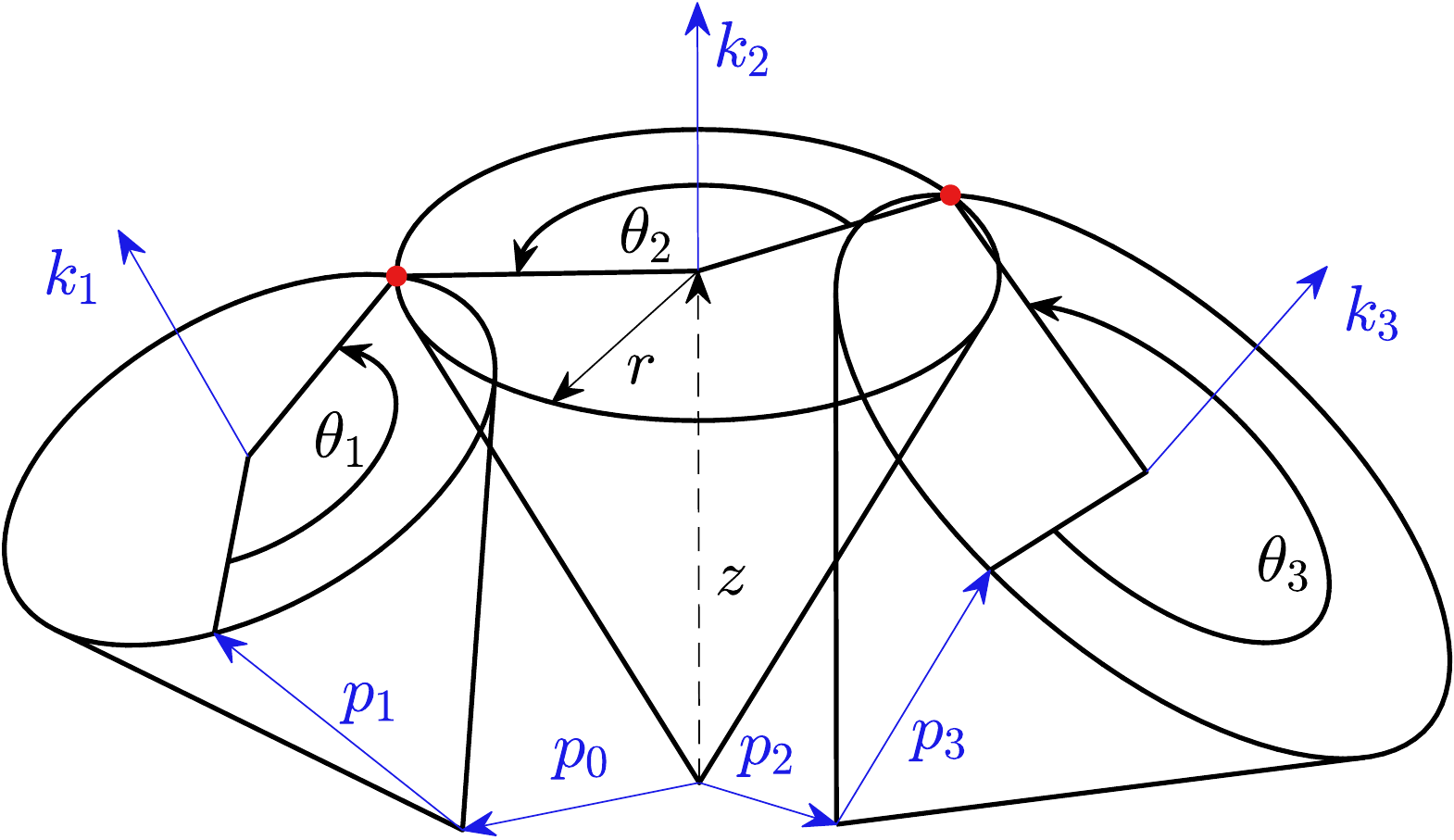}
\caption{Subproblem 5 finds the angles of intersections among three circles, where radius \(r\) and height \(z\) of circle 2 are to be determined and depend on \(\theta_1\) and \(\theta_3\). In the scenario depicted here, there exist other solutions of \((\theta_1, \theta_2, \theta_3)\) associated with different \((r, z)\).}
\label{fig:subprob6pic}
\end{figure}

Equating the two right-hand sides of \eqref{eq:subprob5radius} results in an equation of the form \(P_1 \pm \sqrt{R_1} =P_3 \pm \sqrt{R_3} \), and isolating and squaring for each radical yields \((P_1-P_3)^4 - 2(P_1-P_3)^2(R_1+R_3) + (R_1-R_3)^2=0\), a quartic polynomial in \(z\). (In certain cases, it degenerates to a lower-degree polynomial.) There are many ways to find the roots of a quartic polynomial in closed form \cite{shmakov2011universal}.
For each root \(z_i\), find solutions for \(x_1\) and \(x_3\) by computing the positive and negative branches in \eqref{eq:x1_x3} and checking for equal norm after plugging into \eqref{eq:sp5_circles_linear}. Then, find \(\theta_2\) using Subproblem~1 to solve \eqref{eq:subproblem_5} using the value from \eqref{eq:sp5_circles_linear}. As two ellipses can intersect up to four times, there are up to four solutions for \((\theta_1, \theta_2, \theta_3)\).

\begin{remark}[Coplanar Axes]
The solution procedure simplifies when \(p_0\tr k_1^\cross k_2=0\) or \(p_2\tr k_3^\cross k_2 =0 \), meaning axis 1 or 3 is coplanar with axis 2. In this case, the equation for one of the ellipses degenerates into a line segment, and revolving circles 1 or 3 around \(k_2\) results in a segment of a sphere or plane. If two axes intersect, rewrite the subproblem such that \(p_0=0\) or \(p_2 = 0\) and use Subproblem~3 and then Subproblem~2 to find up to four solutions. If two axes are parallel, use Subproblem~4 and then Subproblem~3 and Subproblem~1.
\end{remark}

\begin{remark}[Decomposition to Subproblem 6]
As was done in \cite{fundamentalsBook}, taking half the squared norm of both sides of \eqref{eq:subproblem_5} yields
\begin{multline}
    p_0 \tr \rot(k_1,\theta_1) p_1 - p_2 \tr \rot(k_3,\theta_3) p_3 \\= \frac{1}{2}(
        \norm{p_2}^2
        + \norm{p_3}^2
        - \norm{p_0}^2
        - \norm{p_1}^2
    ),
\end{multline}
and projecting \eqref{eq:subproblem_5} onto \(k_2\) yields
\begin{equation}
    k_2 \tr \rot(k_1,\theta_1) p_1 - k_2 \tr \rot(k_3,\theta_3) p_3 = k_2 \tr p_2 - k_2 \tr p_0.
\end{equation}
These two equations may be solved with Subproblem~6. Then, \(\theta_2\) can be solved with Subproblem~1.
\end{remark}

% \pagebreak[3]
\begin{subsectionbox}{Subproblem 6: Four Circles}
Given vectors \(p_i\), unit vectors \(k_i\), \(h_i\), \(i=1,\ldots,4\), and scalars \(d_1\), \(d_2\), find $(\theta_1,\theta_2)$
to solve
\begin{equation} \label{eq:subproblem_6}
\begin{cases}
    h_1\tr \rot(k_1, \theta_1)p_1 + h_2\tr \rot(k_2, \theta_2)p_2 = d_1\\
    h_3\tr \rot(k_3, \theta_1)p_3 + h_4\tr \rot(k_4, \theta_2)p_4 = d_2.
\end{cases}
\end{equation}
\end{subsectionbox}
This problem involves four circles defined by \((k_i, p_i)\), where each circle also has a direction vector \(h_i\). The position on the circle gets projected onto the \(h_i\) direction to get a distance for that circle. These circles are coupled so that the sums of distances for circles \((1,2)\) and \((3,4)\) must equal \((d_1, d_2)\), and also coupled so that circles \((1,3)\) and \((2,4)\) have the same rotation angles \((\theta_1, \theta_2)\).

We can use \eqref{eq:rotation_as_linear} to rewrite the subproblem as 
\begin{equation}\label{eq:sp6_linear_eqn}
    \begin{bmatrix}
    h_1\tr A_{k_1,p_1} & h_2\tr A_{k_2,p_2}\\
    h_3\tr A_{k_3,p_3} & h_4\tr A_{k_4,p_4}
    \end{bmatrix}
    \begin{bmatrix}
    x_1\\x_2
    \end{bmatrix}
    = Ax = b =
    \begin{bmatrix}
    b_1\\b_2
    \end{bmatrix},
\end{equation}
where
\begin{subequations}
\begin{align}
    b_1 = d_1 - h_1\tr k_1 k_1\tr p_1 - h_2\tr k_2 k_2\tr p_2,\\
    b_2 = d_2 - h_3\tr k_3 k_3\tr p_3 - h_4\tr k_4 k_4\tr p_4.
\end{align}
\end{subequations}
If row 1 is a multiple of row 2, then \(A\) loses rank and there is a continuum of solutions. In this case, pick any nonzero equation from \eqref{eq:subproblem_6}, choose \(\theta_1\) or \(\theta_2\) arbitrarily (within a valid range), and solve for the other joint angle with Subproblem~4.

If \(A\) is full rank, the unconstrained solutions to \eqref{eq:sp6_linear_eqn} are
\begin{equation}\label{eq:3axes_linear_soln}
    x = x_{min} + \xi_1 x'_{N1} + \xi_2 x'_{N2},
\end{equation}
where \(x_{min} = A^+ b\) is the minimum-norm solution, \(x_N = \xi_1 x'_{N_1} + \xi_2 x'_{N_2}\) is is an arbitrary vector in the two-dimensional null space of \(A\), \(x'_{N_i}\) forms a basis for the null space, and \(\xi_i\) parameterizes the solution. The values of \(x_{min}\), \(x'_{N_1}\), and \(x'_{N_2}\) can be found by QR decomposition of \(A\tr\).

If we impose the constraints \(\norm{x_1}=1\) and \(\norm{x_2}=1\), we get two equations for ellipses in \((\xi_1, \xi_2)\).
Like in Subproblem 5, the intersection of two ellipses can be found by finding the roots of a quartic polynomial. However, in this subproblem the equations for the ellipses are in standard form, so some extra work is needed to find the quartic equation \cite{eberly2000intersection}.
The solutions for \((\xi_1, \xi_2)\) can then be used to find solutions to \((x_1, x_2)\) and therefore the two unknown angles.

As in Subproblem 4, this solution method still works even if \(h_i\) are not unit vectors.

\begin{remark}[Simplified Case]
The problem simplifies if \(h_i\) or \(p_i\) is parallel to \(k_i\) for some \(i\). In this case, one of the equations in \eqref{eq:subproblem_6} reduces to Subproblem~4. After solving for one angle from that equation, the other equation can be solved by using Subproblem~4 again.
\end{remark}
\fi
\section{Polynomial Method Numerical Examples} \label{sec:app_num_examples}
\subsection{Three Pairs of Intersecting Axes: CRX-10iA/L} \label{sec:numerical_CRX}
The FANUC CRX-10iA/L has kinematic parameters
\begin{equation}
\begin{gathered}
    p_{12} = p_{34} = p_{56} = 0,\enspace
    p_{23} = 710 e_z,\\
    p_{45} = 540 e_y + 150 e_z,\\
    h_1 = e_z,\enspace
    h_2 = h_3 = h_5 = e_x,\enspace
    h_4 = h_6 = e_y.
\end{gathered}
\end{equation}
For this example, pick \(R_{06} = I\) and \(p_{06}=0.25 e_x + 0.25 e_y + 0.25 e_z\). We get one equation for the error, one equation for Subproblem~3, and two equations for Subproblem~4:

\begin{subequations}
\begin{multline}
    \frac{2 x_4  {\left({x_1 }^2 -1\right)} {\left(-{x_2 }^2  x_3 -x_2  {x_3 }^2 +x_2 +x_3 \right)}}{{\left({x_1 }^2 +1\right)} {\left({x_2 }^2 +1\right)} {\left({x_3 }^2 +1\right)} {\left({x_4 }^2 +1\right)}}\\
    +\frac{x_1  {\left({x_4 }^2 -1\right)}}{{\left({x_1 }^2 +1\right)} {\left({x_4 }^2 +1\right)}}
    =0,
\end{multline}
\begin{equation}
    \frac{x_1^2 x_4^2  -5 x_1 x_4 ^2 + 4 x_1^2 -4 {x_4 }^2 -5 x_1  -1}{{\left({x_4 }^2 +1\right)}}=0,
\end{equation}
\begin{multline}
\frac{3 {x_2}^2x_4  {\left({x_3 }^2 -1\right)}}{10 {\left({x_3 }^2 +1\right)} {\left({x_4 }^2 +1\right)}}
+\frac{27 {x_2}^2x_3 }{25 {\left({x_3 }^2 +1\right)}}
+\frac{24{x_2}^2}{25}\\
+\frac{27 x_2{\left({x_3 }^2 -1\right)}}{25 {\left({x_3 }^2 +1\right)}}
-\frac{6 x_2x_3  x_4 }{5 {\left({x_3 }^2 +1\right)} {\left({x_4 }^2 +1\right)}}\\
-\frac{3 x_4  {\left({x_3 }^2 -1\right)}}{10 {\left({x_3 }^2 +1\right)} {\left({x_4 }^2 +1\right)}} 
-\frac{27 x_3 }{25 {\left({x_3 }^2 +1\right)}}
-\frac{23}{50}=0,
\end{multline}
\begin{multline}
    \frac{1}{\left({x_4 }^2 +1\right)}
    (6307 {x_3 }^2  {x_4 }^2 +4260 {x_3 }^2x_4 +15336 x_3 {x_4 }^2\\
     + 6307 {x_3 }^2 +6307 {x_4 }^2+15336 x_3\\
      -4260 x_4 +6307)=0.
\end{multline}
\end{subequations}

Clearing the denominators, we get   
\begin{subequations}
\begin{align}
\begin{split}
    P_1 ={}&  x_1x_2^2x_3^2x_4^2  - 2x_1^2x_2^2x_3x_4 - 2x_1^2x_2x_3^2x_4\\
    &- x_1x_2^2x_3^2 + x_1x_2^2x_4^2  + x_1x_3^2x_4^2 + 2x_1^2x_2x_4 \\
    &+ 2x_1^2x_3x_4  + 2x_2^2x_3x_4   + 2x_2x_3^2x_4  - x_1x_2^2  \\
    &- x_1x_3^2  + x_1x_4^2 - 2x_2x_4 - 2x_3x_4 - x_1  = 0,
\end{split}\\[1em]
    P_2 ={}& x_1 ^2 x_4 ^2 -5x_1 x_4 ^2 +4x_1 ^2  -4x_4 ^2 -5x_1 -1 = 0, \\[1em]
\begin{split}
    P_3 ={}& 48x_2^2x_3^2x_4^2 + 54x_2^2x_3x_4^2 + 15x_2^2x_3^2x_4 \\
    &+ 54x_2x_3^2x_4^2 + 48x_2^2x_3^2  + 48x_2^2x_4^2 - 23x_3^2x_4^2 \\
    &+ 54x_2^2x_3  - 15x_2^2x_4  + 54x_2x_3^2 - 60x_2x_3x_4  \\
    &- 54x_2x_4^2 - 15x_3^2x_4 - 54x_3x_4^2 + 48x_2^2 - 23x_3^2 \\
    &- 23x_4^2  - 54x_2 - 54x_3 + 15x_4 - 23 = 0,
\end{split}\\[1em]
\begin{split}
     P_4 ={}& 6307x_3^2x_4^2 + 4260x_3^2x_4 + 15336x_3x_4^2 + 6307x_3^2 \\
     &+ 6307x_4^2 + 15336x_3  - 4260x_4 + 6307 = 0.
\end{split}
\end{align}
\end{subequations}
These equations are highly decoupled.
\(P_1\) depends on \((x_1, x_2, x_3, x_4)\),
\(P_2\) on \((x_1, x_4)\),
\(P_3\) on \((x_2, x_3, x_4)\), and
\(P_4\) on \((x_3, x_4)\). We can eliminate three variables by only taking resultants three times. First, we combine \((P_1, P_2)\) to eliminate \(x_1\). Then, we eliminate \(x_2\) by combining with \(P_3\). Finally, we eliminate \(x_3\) by combining with \(P_4\). The factor of interest is the order 16 polynomial in \(x_4\):
\begin{multline}
    592628990976x_4^{16} - 493649640000x_4^{15}\\
    - 7228671199758x_4^{14} - 8539013655000x_4^{13}\\
    - 29945708282222x_4^{12} + 2896653453750x_4^{11}\\
    + 45929194539844x_4^{10} + 39808558518750x_4^9\\
    + 177776490417945x_4^8 - 39808558518750x_4^7\\
    + 45929194539844x_4^6 - 2896653453750x_4^5\\
    - 29945708282222x_4^4 + 8539013655000x_4^3\\
    - 7228671199758x_4^2 + 493649640000x_4\\
    + 592628990976 = 0.
\end{multline}
The eight real roots of this polynomial correspond to the eight IK solutions for this pose.

\subsection{Two Intersecting Axes: RRC with Fixed Joint 6} \label{sec:numerical_RRC}
We fix \(q_3 = \pi/2\) and choose the pose
\begin{subequations}
\begin{align}
R_{06} &= \rot(e_y, \pi/6) \rot(e_z,\pi/9) \rot(e_y, \pi/18),\\
p_{06} &= 50.12 e_x -355.34 e_y +  354.56 e_z,
\end{align}
\end{subequations}
to find three equations in \((x_1, x_3, x_4)\), where one equation has 200 terms and two equations each have 16 terms. We eliminate \(x_1\) to get two equations with 241 and 10 terms. Then, we eliminate \(x_3\) to get one equation with 75 terms. Factoring, we get a polynomial of degree 10 with eight real solutions for \(x_4\).

\subsection{General 6R: Husty et al. Robot} \label{sec:numerical_husty}
We will apply the polynomial method to a highly general robot described in \cite{husty2007new}. The nominal kinematic parameters are
\begin{equation}
\begin{gathered}
    p_{12} = 10 e_x + 100 e_z,\ 
    p_{23} = 100 e_x + 25 e_y + 25 \sqrt{3} e_z,\\
    p_{34} = 150 e_x + 50 e_z,\ 
    p_{45} = 20 e_x + 25 \sqrt{3} e_y - 25 e_z,\\
    p_{56} = 5 e_z + 10 e_y - 3 \sqrt{3} ez,\\
    h_1 = h_3 = h_6 = ez,\ 
    h_2 = 1/2 e_y + \sqrt{3}/2 ez,\\
    h_4 = -\sqrt{3}/2 e_y + 1/2 e_z,\ 
    h_5 = -1/2 e_y + \sqrt{3}/2 e_z.
\end{gathered}
\end{equation}
Applying the rational approximation with \(\theta = \pi/6\) gives \(\tilde c = 564719/652081\), \(\tilde s = 326040/652081\), and the approximate kinematic parameters
\begin{multline}
    \tilde h_2 =  \tilde s e_y + \tilde c e_z,\ 
    \tilde h_4 = -\tilde c e_y + \tilde s e_z,\ 
    \tilde h_5 = -\tilde s e_y + \tilde c e_z.
\end{multline}
We can approximate the position vectors in a similar way:
\begin{subequations}
\begin{align}
    \tilde p_{23} &= 100 e_x + 25 e_y + 23859/551 e_z,\\
    \tilde p_{45} &= 20 e_x + 23859/551 e_y - 25 e_z,\\
    \tilde p_{56} &= 5 e_z + 10 e_y -46367/2677 ez.
\end{align}
\end{subequations}

The goal is to find all IK solutions corresponding to the end effector pose given by the forward kinematics (based on the nominal kinematics parameters) of the joint angles
\begin{equation}
    q = \begin{bmatrix}
    -\frac{\pi}{6} & \frac{\pi}{2} & -\frac{\pi}{3} & \frac{\pi}{2} &  \frac{\pi}{6} & -\frac{\pi}{6}
    \end{bmatrix}\tr.
\end{equation}

The desired pose is approximated as
\begin{equation}
\begin{gathered}
    R_{06} = \rot(e_z, \gamma)\rot(e_y, \beta)\rot(e_z, \alpha),\\
    \tan( \tilde \beta /2 ) = 571/1284,\ 
    \tan( \tilde \gamma /2 ) = -2858/871,\\
    \tilde p_{06} = 73288/345 e_x
        + 33409/245 e_y
        + 655/97 e_z.
\end{gathered}
\end{equation}
Applying the subproblem decomposition, we have four polynomials \((P_1, P_2, P_3, P_4)\) in four unknowns, \((x_1, x_2, x_3, x_5)\), with \((189, 54, 62, 62)\) terms, respectively. As \(P_2\) has the smallest number of terms, we take the resultant of \(P_2\) with the remaining polynomials while eliminating \(x_5\) to get three polynomials \((R_1, R_2, R_3)\), where we only keep factors which depend on \((x_1, x_2, x_3)\). We combine \((R_1, R_3)\) while eliminating \(x_3\) and get a product of a polynomials with degrees \((8, 8, 48, 96)\) which depend on \((x_1, x_2)\). We keep the polynomial of degree 48, which we call \(V_1\). We also combine \((R_2, R_3)\) while eliminating \(x_3\) and get a product of polynomials with degrees \((32, 64)\) which depend on \((x_1, x_2)\), and keep the polynomial with degree 32, which we call \(V_2\). We combine \((V_1, V_2)\) and get a product of polynomials with degrees \((2, 4, 4, 4, 4, 16, 32, 56, 536)\) and of multiplicity \((16, 4, 8, 8, 4, 1, 1, 1, 1)\), respectively. The solutions for \(x_1\) correspond to the zeros of the 16-degree polynomial (with multiplicity one).

To find \(x_2\), we plug each solution of \(x_1\) into \(V_1\) and \(V_2\) and find any common zeros.
The remaining joint angles are found in closed form using subproblems. By also including any complex zeros, we find all 16 solutions for \(q\) to higher precision than shown in \cite{husty2007new}.

\subsection{General 6R: YuMi with Fixed Joint 3} \label{sec:numerical_yumi}
For this 7R robot, we fix \(q_3 = \pi/2\) and choose \(R_{06} = I\) and \(p_{06} = 50e_x-355e_y+354e_z\). Staring with four polynomials with \((10, 10, 46, 56)\) terms, we eliminate three variables. The solutions for \(x_1\) are the eight real zeros of an order 12 polynomial.

\section*{Acknowledgment}
\addcontentsline{toc}{section}{Acknowledgment}
The authors would like to thank C. Carignan, D. Carabis, and A. Gostin for their fruitful discussions. The authors would also like to thank R.~Chen, A.~Maksumi\`{c}, J.~Mathew, and A.~Ropp for their careful work in implementing the subproblem and IK solutions. Finally, the authors would like to thank the reviewers of this paper for their comprehensive comments and recommendations.

\IEEEtriggeratref{66} % Insert a column break to equalize last 2 columns
% \nocite{*} % List out whole bib to check for anything uncited 
\bibliographystyle{IEEEtran}
\bibliography{subproblem_survey, bib, robot_kin_refs}
\end{document}